\documentclass{article}

    \usepackage[preprint]{neurips_2023}

\usepackage[utf8]{inputenc} % allow utf-8 input
\usepackage[T1]{fontenc}    % use 8-bit T1 fonts
\usepackage{hyperref}       % hyperlinks
\usepackage{url}            % simple URL typesetting
\usepackage{booktabs}       % professional-quality tables
\usepackage{amsfonts}       % blackboard math symbols
\usepackage{nicefrac}       % compact symbols for 1/2, etc.
\usepackage{microtype}      % microtypography
\usepackage{xcolor}         % colors
\usepackage{colortbl}
\usepackage{wrapfig}
\usepackage{graphicx}
\usepackage{lipsum}
\usepackage{multirow}
\usepackage{textcomp}
\usepackage{caption}
\usepackage{subcaption}
\usepackage{amsmath}
\usepackage{natbib}
\usepackage{ifthen}
\newboolean{review_flag}
\setboolean{review_flag}{false}

\setcitestyle{numbers,square}

\title{{RPTQ}: Reorder-based Post-training Quantization for Large Language Models}

\author{%
  Zhihang Yuan\thanks{\, Equal contribution. hahnyuan@gmail.com, linniu@hust.edu.cn. This work was done when Lin Niu and Jiawei Liu were interns at Houmo AI. $\dag$\,  Corresponding author. xgwang@hust.edu.cn, bingzhewu@tencent.com. }\\
  Houmo AI \\
  \And
  Lin Niu$^{*}$\quad Jiawei Liu\quad Wenyu Liu\quad Xinggang Wang$^{\dag}$\\
  Huazhong University of Science \& Technology
  \And
  Yuzhang Shang \\
  Illinois Institute of Technology \\
  \And
  Guangyu Sun \\
  Peking University \\
  \And
  Qiang Wu\\
  Houmo AI \\
  \And
  Jiaxiang Wu\\
  Tencent AI Lab
  \And
  Bingzhe Wu$^{\dag}$ \\
  Tencent AI Lab \\
}

\begin{document}

\maketitle

\begin{abstract}
Large-scale language models (LLMs) have demonstrated impressive performance, but their deployment presents challenges due to their significant memory usage. This issue can be alleviated through quantization.
In this paper, we identify that the challenge in quantizing activations in LLMs arises from varying ranges across channels, rather than solely the presence of outliers.
To address this challenge, we introduce a quantization method called RPTQ, which utilizes a reorder-based approach. 
By rearranging the channels and quantizing them in clusters, RPTQ effectively mitigates the impact of range differences between channels.
To minimize the overhead of the reorder operation, we fuse it into the layer norm operation and weights in linear layers. 
In our experiments, RPTQ achieved a significant breakthrough by utilizing 3-bit activation in LLMs for the first time, resulting in a substantial reduction in memory usage.
For instance, quantizing OPT-175b can lead to a memory consumption reduction of up to 80\%.
% \syz{Maybe we can add a more straightforward achievement, for example, make OPT175B from 8 A6000 to one by saving the memory cost.}
  \ifthenelse{\boolean{review_flag}}{}{The code is in \href{https://github.com/hahnyuan/RPTQ4LLM}{https://github.com/hahnyuan/RPTQ4LLM}.}
  % \syz{I think codebase linking to authors should not be attached when submitting}
  % demonstrating the potential for significant reductions in memory and computation requirements for large-scale language models while maintaining high levels of accuracy. 
\end{abstract}
\section{Introduction}

% LLM取得了很好的成绩，但是LLM模型太大了，阻碍了它的发展。
% 具体来说，权重和激活值都会占用很大的空间。如果没有足够的显存，将会来回搬数，很慢。
% 举个例子，在OPT-175b，如果使用FP16的数据存储权重和激活值，将会有xxx的内存空间占用。一般的gpu放不下。

% 量化是对网络进行压缩的
Large-scale language models (LLMs) have demonstrated impressive performance in various tasks, but their deployment poses challenges due to their enormous model size.
For example, the OPT-175B model~\cite{zhang2022opt} contains 175 billion parameters, which require significant memory to store
As the sequence length and batch size increase, the problem of memory consumption becomes more severe because activations.
In some cases, the key and value cache can consume more than 100 times the memory of the weights.
However, a single GPU or server does not possess sufficient memory capacity to store such massive weights and activations. 
To address this issue, LLMs are often divided into multiple chunks and stored on different devices. 
However, this requires data to be transferred between devices during computation, leading to significant bandwidth and energy consumption~\cite{aminabadi2022deepspeed,sheng2023high}.

% Large-scale language models (LLMs) have demonstrated exceptional performance on various tasks, but their deployment poses challenges due to their enormous model size.
% The OPT-66B model~\cite{zhang2022opt}, for example, contains 66 billion parameters, which requires significant memory and energy resources to store and run. 
% % Storing the weights and activation values (such as the key and value cache) of LLaMA-66B using floating-point 16 (FP16) format can result in the consumption of nearly 200GB of storage space. 
% When the sequence length and batch size increase, the problem in memory consumption becomes more severe because the activation also takes memory.
% However, a single GPU or server does not have sufficient memory capacity to store such large amounts of model weights and intermediate activations. 
% As a result, the LLM model needs to be partitioned into multiple chunks and stored in different devices. 
% Since the weights and activations are stored in different devices, data must be transferred between these devices during computation, which can result in significant bandwidth and energy consumption.
% Therefore, efficient storage and computation strategies are crucial for improving the efficiency and scalability of large-scale language models.
% 如果权重和激活值使用FP16来存储，总共会占用接近200GB的存储空间。单块GPU没有那么大的显存可以存储如此大的模型的权重和中间激活值，因此需要将模型分块来进行计算。数据在多个存储设备之间的传输将成为瓶颈。

To address the challenges posed by LLMs' high memory usage, model quantization has emerged as a promising solution. 
This technique involves quantizing both the weights and activations of LLMs using low-bit integers, resulting in a significant reduction in storage and computational costs. 
Specifically, quantization reduces memory requirements for saving weights and activations and accelerates compute-intensive operations like Matrix Multiplication and linear layers.
By quantizing weights and activations, storage and communication overhead is reduced, leading to improved efficiency and faster inference times.
Quantization methods are typically divided into two categories: post-training quantization (PTQ) and quantization-aware training (QAT). 
While QAT methods can lead to higher accuracy in most cases, they require significant computational resources to train the models, making them less practical for LLMs that already have significant training costs. 
In contrast, PTQ methods can quantize pre-trained models without additional training, making them more practical for larger models that require significant computational and memory resources. 
This paper focuses on PTQ for LLMs.
% Our approach leverages the benefits of PTQ by proposing an accurate and efficient PTQ solution.
% Overall, quantization provides an effective solution for addressing the computational and memory constraints associated with deploying large-scale language models.

\begin{figure}[tb] % h:here 当前位置 % b bottom % t top % p 浮动
    \centering
    \includegraphics[width=0.96\textwidth]{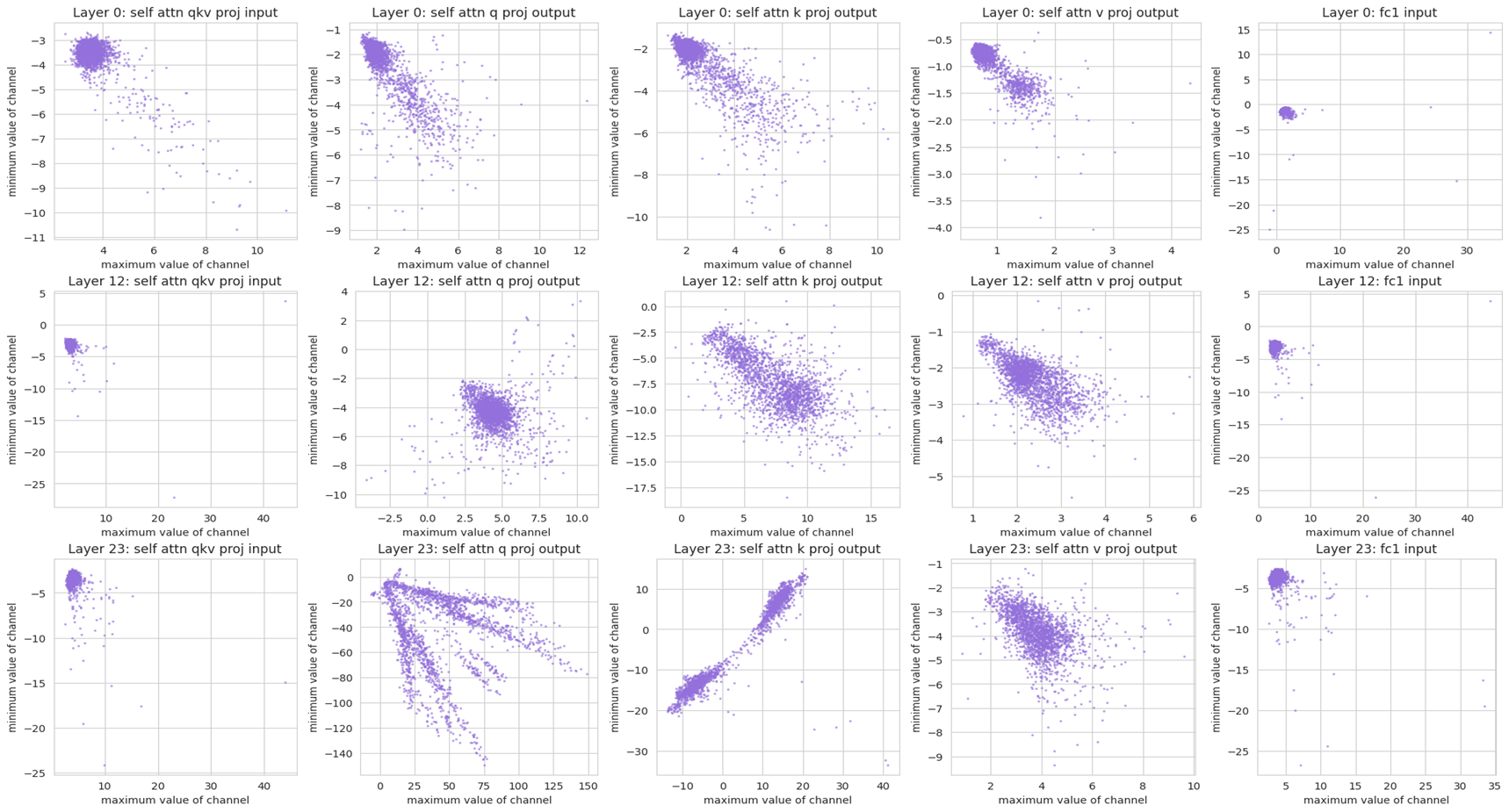} %ims/xx.png
    \caption{Demonstration of the distribution of different channels in OPT decoder layers. Each point is (maximum value, minimum value) of a channel in the activation.}
    \vspace{-10pt}
    \label{im_distribution}
\end{figure}

In this paper, we highlights the challenge of quantizing the activations of LLMs, which is attributed to the significant variations in the values across different channels~\footnote{For simplicity, we use the term "channel" to refer to the dimension of the hidden size. \ifthenelse{\boolean{review_flag}}{}{See Appendix~\ref{appendix_distribution} for more results.}}, as shown in Figure~\ref{im_distribution}. 
Two observations can be made from this figure: 1) Some channels exhibit significant outliers, with maximum or minimum values that are hundreds of times larger than those of other channels. 
Previous studies~\cite{xiao2022smoothquant,dettmers2022llm} have also identified this issue and proposed special treatment for outliers. 
2) Different channels exhibit significant difference in the range of values. Quantizing different channels using the same quantization parameter can lead to substantial quantization errors. 
Even if two channels have the same absolute value of outliers, they can exhibit strong difference in the range of numerical values. 
For instance, one channel may have a range of -100 to -50, while another channel may have a range of 80 to 100.
Using the same quantization parameters for them can lead to significant quantization errors, which is a challenge that has not been effectively addressed in previous works.
To address the issue of quantizing activations with channels that have significantly different ranges, we propose a method called RPTQ.
This method involves clustering channels in activations that exhibit similar value ranges, followed by the quantization with the same quantization parameter to the values within each cluster.
Consequently, channels displaying considerable discrepancies in numerical ranges can utilize distinct quantization parameters, leading to a significant reduction in quantization error.
% We cluster the channels in activations with similar value ranges together. 
% We then quantize values in each cluster with the same quantization parameter.
% This approach enables channels with significant differences in numerical range to use distinct quantization parameters, resulting in a substantial reduction of quantization error.
Furthermore, we propose strategies to avoid explicit reordering, thereby decreasing computational overhead and enhancing inference efficiency. 
We propose a modified layer norm operation to yield reordered activations directly, obviating the necessity for explicit channel adjustments during the inference process.
In addition, we reorganize the weights of linear layers to enable them to directly accept and produce activations in a sorted order.

Our experiments demonstrate that RTPQ is an effective solution for addressing the issue of quantizing the activations of LLMs. 
Clustering the channels in only a small number of clusters can significantly reduce quantization errors and improve the accuracy of quantized LLMs. 
% The results show that we achieved a breakthrough by pushing LLM models to 3-bit activation for the first time. 
The results show that RPTQ can achieve significant reductions in memory for LLMs while maintaining high levels of accuracy. 
For instance, by quantizing OPT-175b, memory usage can be reduced by 73\% with a perplexity loss of less than 0.5 or by 80\% with a perplexity loss of less than 1.5 across different sequence lengths and batch sizes.

% Overall, our proposed approach represents a promising step forward in the development of efficient and accurate large-scale language models.

\section{Related Work}

\subsection{Large Language Model}

Large Language Models (LLMs) have shown immense potential for various applications. 
A lot of LLMs have been developed~\cite{devlin2018bert, brown2020language, scao2022bloom, zeng2022glm, black2022gpt, touvron2023llama}. 
These models have exhibited exceptional performance, but at the cost of significant memory consumption, which poses significant challenges for their deployment~\cite{li2022large}.
Ways to solve this problem include model compression~\cite{frantar2023massive}, distributed computing~\cite{aminabadi2022deepspeed} and computational graph optimization~\cite{dao2022flashattention}.
In this study, we focus on compressing LLMs through quantization.
% the activation of OPT~\cite{zhang2022opt}. Further investigations on other networks will be conducted in future research.

\subsection{Quantization}

Quantization is an important technique for reducing the computational and memory requirements of deep neural networks (DNNs). 
There are two main categories of quantization methods for DNNs: post-training quantization (PTQ)~\cite{zhao2019improving, wang2020towards, nagel2020up, li2021brecq, yuan2022ptq4vit} and quantization-aware training (QAT)~\cite{courbariaux2016binarized,zhou2016dorefa, choi2018pact, jung2019learning}. 
PTQ methods involve quantizing pre-trained models, while QAT methods involve training models with quantization constraints.

While QAT methods have been shown to improve the accuracy of DNNs in some cases, they require significant computational resources to train the models. 
For instance, LSQ introduces a differentiable quantization function, which enables gradient-based optimization during training~\cite{esser2019learned}. 
LSQ involves quantizing and de-quantizing the activation and weights, which requires additional computations. 
Additionally, LSQ involves optimizing the quantization parameters, which requires extra training epochs to solve and update the gradient of the parameters.
This makes them less practical for large-scale language models (LLMs) that already have high training costs. 
In contrast, PTQ methods are more feasible for LLMs, as they involve quantizing pre-trained models, which do not require additional training time.

% 最近一年来，有一些为LLM设计的量化方法，比如SmoothQuant，GPTQ，还有LLM.int8()。
% ZeroQuant-LLM.int8()-smoothquant-gptq
Recently, there are some multi-billion scale transformer quantization methods designed for LLMs.
ZeroQuant~\cite{yao2022zeroquant} proposes a fine-grained quantization scheme that can be applied to both weights and activations. 
It treats each layer of the transformer as a small neural network and uses the FP model to distill the quantization model.
nuQmm~\cite{park2022nuqmm} utilizes group-wise binary-coding non-uniform quantization scheme, and propose a specialized multiplicative kernel to speed up the operation.
LLM.int8()~\cite{dettmers2022llm} observes that a significant contributor to poor quantization performance is outliers in activations.
The method fixes this with mixed-precision quantization.
SmoothQuant~\cite{xiao2022smoothquant} migrates the quantization difficulty from activations to weights by proposing a mathematically equivalent per-channel scaling transformation. 
This transformation smooths the magnitude across channels, making the model quantization-friendly.
GPTQ~\cite{frantar2022gptq} uses second-order approximation to quantize weights, enabling the weight quantization of LLMs into 4-bit - the first post-training method to do so. 
However, these methods can only achieve the quantization of activations to 8 bits.
Comprehensive study~\cite{yao2023comprehensive} has improved ZeroQuant, treating each linear layer of the transformer as a small neural network for distillation, and achieved usable performance at W4A8 quantization.

PTQ-SL~\cite{yuan2021ptq} proposed that adjusting the channel order of weights can lead to higher accuracy in finely-quantized networks. 
However, PTQ-SL mainly focuses on the quantization of weights in convolutional networks, and does not address the quantization issues of activations. 
PGQ~\cite{bondarenko2021understanding} employs a range-based permutation of the embedding dimensions and share quantization parameters among elements in the same group to address the problem of activation quantization.
Nonetheless, it only consider for the dynamic range and utilizes uniformly divided groups, rendering it less efficacious for LLMs.
% \syz{I think we should add one sentence highlighting RPTQ, for example: Our approach, RPTQ, is the first work advancing LLM PTQ into 4W4A (and even 3W3A).}
% In our work, we focus on channel adjustment for activations in LLM networks due to the significant differences among different channels.
% PTQ-SL提出通过调整weight的通道顺序，可以让细粒度的量化网络达到更高的精度。
% 然而，PTQ-SL主要针对卷积网络中的weight，没有考虑activation的量化问题。
% 本文中，我们主要针对LLM网络中，activation不同通道之间的差异来进行通道调整。
\section{PTQ on LLM}

\subsection{Post-training Quantization}

Post-training quantization is a powerful technique for compressing neural networks. 
Although non-uniform quantization can achieve a relatively small quantization error, they require specialized hardware that is not widely accessible~\cite{guo2022ant}. 
In contrast, uniform quantization is a more practical and feasible approach that can be efficiently executed on regular hardware devices. 
Therefore, our study focuses on uniform quantization techniques.
Typically, we use uniform quantization function $Q_k$ to transform a float value $x$ to $k$ bits integer $x_q$ as follows:
\begin{equation}
    x_q = Q_k(x,s,z)=\text{clamp}(\text{round}(\frac{x}{s})+z,-2^{k-1},2^{k-1}-1),
\label{quantization_function}
\end{equation}
where $s$ represents the scaling factor, $z$ denotes the zero point, and the clamp function constrains the value within the range of a $k$-bit integer, specifically $[-2^{k-1},2^{k-1}-1]$. 
For a 4-bit integer, the range is [-8, 7]. The integer $x_q$ can be de-quantized to $\hat{x}=s(x_q-z)\approx x$.
The de-quantized value $\hat{x}$ is a float. 
The quantization parameters, scale factor $s$, and zero point $z$ must be stored in memory for both quantization and de-quantization processes.
To further reduce the storage and computational overhead of quantization, multiple weights or activation values $X=\{x_1,...,x_n\}$ share the same quantization parameters.

There are three steps in post-training quantization (PTQ).
The first step is to specify the quantization settings, which include the bit-width $k$ and quantization type.
The bit-width determines the number of bits used to represent a numerical value in a quantized format. 
The quantization types include static and dynamic quantization. 
Static quantization is a method in which the quantization parameters of activations are computed prior to deployment, and all inputs during runtime are quantized using the same set of parameters. 
% This approach leads to faster inference times. 
% However, since different inputs may have different dynamic ranges of activations, the pre-computed quantization parameters may not be optimal for some inputs, which can lead to a reduction in accuracy.
% Static quantization calculate the quantization parameters of activations before deployment. 
% 在部署后运行过程中，所有的输入都将使用相同的量化参数来对激活值进行量化。
% This method leads to faster inference times. 
% However, 不同的输入可能会有不同的激活值的动态范围，提前设置好的量化参数可能不适合某些输入时的量化, which can result in a drop in accuracy.
Dynamic quantization, on the other hand, set the quantization parameters during runtime~\cite{yao2022zeroquant}. 
% The quantization parameters are computed dynamically for each input data.
% For example, 一种常用的方法是MinMax动态量化，it compute the max value $X_{max}=max(X)$ and min value $X_{min}=min(X)$.
% Then it set the $s=\frac{X_{max}-X_{min}}{2^{k}}$ and $z=round(\frac{X_{max}+X_{min}}{2\times s})$.
Dynamic quantization is generally more accurate than static quantization, as it can adapt to the distribution of activations. 
However, it can result in slower inference times, as the quantization parameters should be computed on-the-fly for each input data.
We focus on static quantization in this paper.
% The choice between static and dynamic quantization depends on the specific application and the trade-off between accuracy and speed.

Next, a calibration dataset consisting of input samples is used to compute the activations of each layer.
The purpose of this step is to capture the distribution of numerical values that are likely to appear during inference. 
Finally, the quantization parameters for activations and weights are selected.
There are many methods to select the parameters.
One of the commonly used methods is Min-Max method. 
This method involves computing the maximum value $X_{max}=max(X)$ and minimum value $X_{min}=min(X)$ of tensor $X$ that share the quantization parameters. 
The scaling factor $s$ and zero point $z$ are set as follows:
\begin{equation}
    s=\frac{X_{max}-X_{min}}{2^{k}},\quad z=-round(\frac{X_{max}+X_{min}}{2 s}).
\label{eq_minmax_quant}
\end{equation}
% We should select the quantization parameters to minimize the quantization error. 
% For instance, in the case of simple tensor-wise quantization, the optimization objective can be expressed as follows:
% \begin{equation}
%     \mathop{\mathrm{argmin}}_{s_W,z_W} L(W,\hat{W}),
% \label{eq_dynamic_quant}
% \end{equation}
% where $W$ is the original weight and $\hat{W}$ is the simulated weight after quantization with scale $s_W$ and zero point $z_W$.
% $L$ is a function used to compute the distance between the output before and after quantization, which is commonly measured using Mean Squared Error (MSE).

\subsection{Challenges in Activation Quantization}

\begin{figure}[tb] % h:here 当前位置 % b bottom % t top % p 浮动
    \centering
    \includegraphics[width=0.85\textwidth]{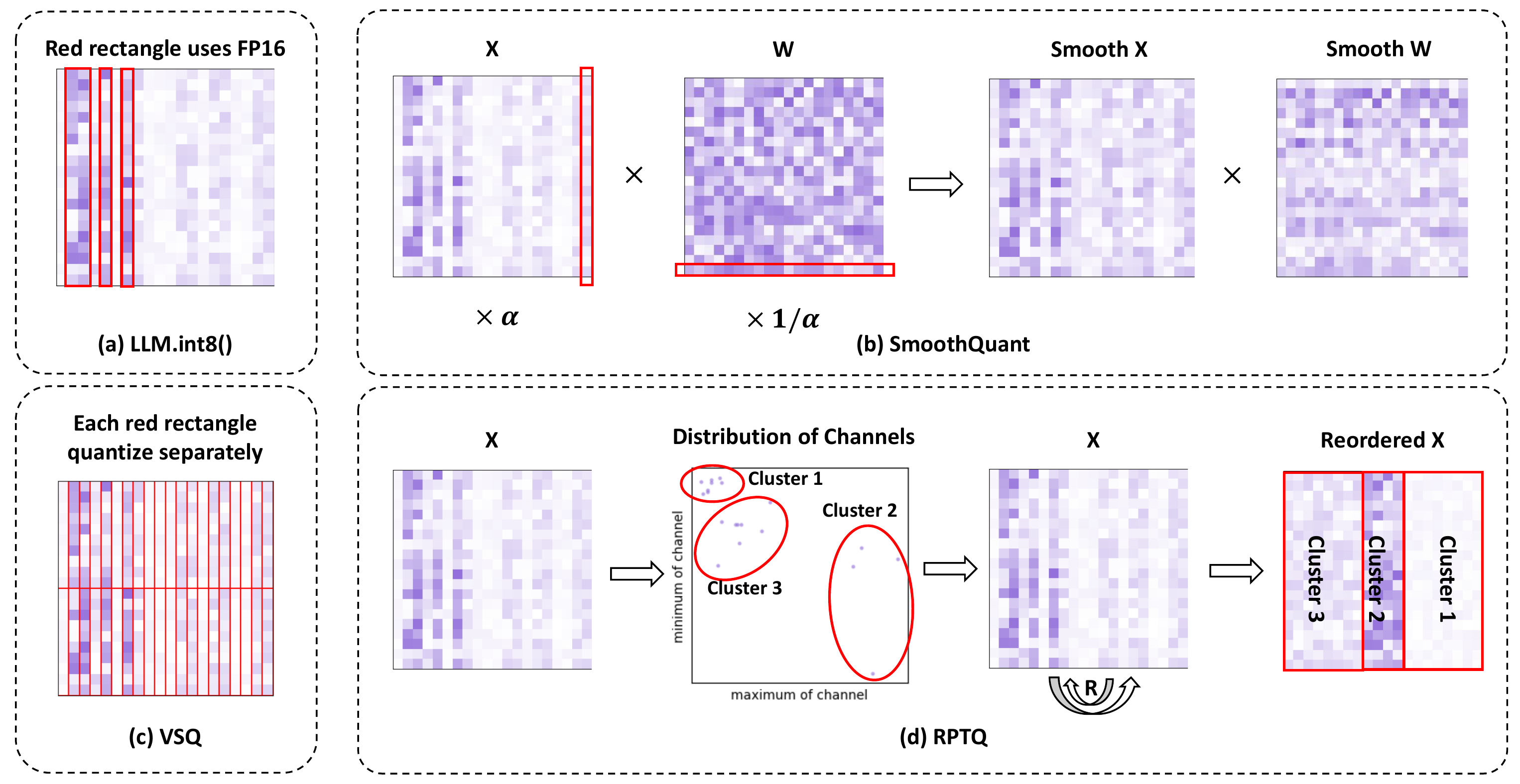} %ims/xx.png
    \caption{Demonstration of different methods to address the problem in quantizing activations.
    }
    \label{im_comparing}
\end{figure}

Recently, the weights of LLMs have been successfully quantized to 4 bits or even lower~\cite{frantar2022gptq} with PTQ. 
However, quantizing the activations in LLMs remains a challenging task.
As shown in Figure~\ref{im_distribution}, the activations in LLMs exhibit significant variations in value range across different channels, which can have a significant impact on the quantization process. 
Per-tensor quantization techniques, which quantize the entire tensor using the same quantization parameters, may not be effective. 
The reason is setting the quantization range to cover a large value range may result in channels with small numerical values taking a large quantization error, while setting it to cover a small value range may lead to significant truncation of outliers and resulting in significant quantization errors.
For instance, one channel may have a range of -100 to -50, while another channel may have a range of 80 to 100. 
Attempting to cover their ranges by quantizing from -100 to 100 would result in a significant quantization error for both channels.

% As shown in Figure~\ref{im_distribution}, the activations in LLMs exhibit significant variations in the magnitude of the numerical values across different channel.
% This variation can have a significant impact on the quantization process. 
% When quantizing the activations of LLMs, traditional per-tensor quantization techniques, which quantize the entire tensor by the same quantization parameters, may not be effective.
% If the quantization is set to cover a large value range, this can result in the channels with small numerical values takes a large quantization error. 
% On the other hand, if the quantization is set to cover a small value range, the outliers will be significantly truncated, also leading to significant quantization errors. 
% As a result, traditional quantization techniques that work well for other types of models may not be effective for LLMs.

Previous research has proposed several methods to address the issue of quantizing activations in LLMs. 
As shown in Figure~\ref{im_comparing}(a), LLM.int8()\cite{dettmers2022llm} utilizes mixed-precision quantization by using high-precision data types (FP16) to quantize the outliers in activations and low-precision data types (INT8) for the remaining values. 
Using FP16 for these few exceptional channels can prevent errors caused by quantization.
As shown in Figure~\ref{im_comparing}(b), SmoothQuant\cite{xiao2022smoothquant} tackles the quantization difficulty by introducing a mathematically equivalent per-channel scaling transformation that smooths the magnitude across channels, making the activations more amenable to quantization. Specifically, the activation channels with large values are multiplied by a small factor $\alpha$, and the corresponding weights for processing these channels in the next layer are multiplied by $1/\alpha$.
However, SmoothQuant may not be effective in addressing the high variance of value range across channels.
Additionally, this approach may also lead to issues with weight quantization.
As shown in Figure~\ref{im_comparing}(c), Per-vector scaled quantization (VSQ)~\cite{dai2021vsq,keller202217vsq,dettmers2022llm}, has been proposed sharing quantization parameters for adjacent columns or rows of n activation values.
Although the approach of setting a quantization parameter for adjacent n values can help alleviate the issue of varying numerical values, the computation and storage overhead of calculating and storing these parameters for each set can be significant.

\section{Reorder-based Quantization}

\subsection{Clustering and Reordering of Channels}

In the section above, it was observed that there are notable variations in the activations across channels.
To solve this problem, we propose a novel reorder-based quantization approach called RPTQ. 
The main idea of this approach is to cluster the channels in the activations and reorganize them for quantization as shown in Figure~\ref{im_comparing}(d).

To implement our reorder-based quantization approach, we first utilize a calibration dataset as input for inference, from which we derive the maximum and minimum values for each activation channel. 
Subsequently, we employ the K-Means algorithm~\cite{macqueen1967classification} to categorize the distinct channels into $g$ clusters, based on the points formed by each channel's maximum and minimum values.
Once the clusters are established, we proceed with channel reordering by positioning channels from the same cluster together. 
Following the reordering process, we quantize the activations within each cluster.
Specifically, we calculate the quantization parameters (scale $s$ and zero point $z$) individually for each cluster. 
As a result, channels with analogous maximum and minimum values are assembled together and share the same quantization parameters. 
This method guarantees optimization of the quantization process for every cluster, ultimately reducing the quantization error.

% To implement our reorder-based quantization approach, we first use a calibration dataset as input to inference, and obtain the maximum and minimum values for each channel in the activations.
% Next, we use the KMeans algorithm~\cite{macqueen1967classification} to cluster the different channels into $g$ clusters based on the points formed by the maximum and minimum values of each channel.
% Once we have obtained the clusters, we perform channel reordering, placing channels from the same cluster together. 
% In this way, channels with similar maximum and minimum values are grouped together, and they share the same set of quantization parameters.
% After reordering, we quantize the activations in each cluster. 
% Specifically, we compute the quantization parameters (scale $s$ and zero point $z$) for each cluster separately, ensuring that the quantization parameters are tailored to the specific channel. 
% This approach ensures that the quantization process is optimized for each cluster and reduces the quantization error.

We formalize the reordering process as follows:
Let $X \in \mathbb{R}^{B\times N \times C}$ be the activation tensor, where $B$ is the number of calibration samples, $C$ is the number of channels and $N$ is the number of tokens.
We first compute the minimum and maximum values of each channel, denoted as $X_{\min} \in \mathbb{R}^C$ and $X_{\max} \in \mathbb{R}^C$, respectively, using the calibration dataset:
\begin{equation}
    X_{\min}=\min_{n=1}^N\min_{b=1}^B X_{b,n}, \quad
    X_{\max}=\max_{n=1}^N\max_{b=1}^B X_{b,n}. 
\end{equation}
Then, we group the channels into $g$ clusters using K-means clustering based on the values of $(X_{\min,i}, X_{\max,i})$ for each channel.
Let $S^1, S^2, ..., S^g$ be the sets of channels' indexes in each cluster, where $S^i \subseteq \{1, 2, ..., C\}$ and $\bigcup_{i=1}^{g}S^i = \{1, 2, ..., C\}$.
Finally, we reorder the channels in $X$ based on the indexes.
% Let $S^i = [s_1^i, s_2^i, ..., s_{n_i}^i]$ denote the set of indices obtained by ordering the channels in cluster $i$, where $n_i$ is the number of channels in cluster $i$. 
We concatenates all the indices as a vector $S=[S^1, S^2, ..., S^g]$.
We obtain the reordered activation tensor $\tilde{X}_{:,:,i}=X_{:,:,S_i}$.
% The weights are also reordered accordingly to maintain consistency with the reordered activations.

As illustrated in Figure~\ref{im_comparing}, our approach presents several advantages compared to previous methods. 
Firstly, RPTQ is more adept at addressing the challenge of channel difference in activations. 
By clustering channels with similar value ranges, it diminishes the influence of both outliers and range differences. 
Secondly, RPTQ exhibits memory and computation efficiency, as it only requires managing quantization parameters for each cluster rather than each individual channel or vector. 

% To be specific, 
% 我们将校准数据集作为输入，运行网络以获得activation每一个通道的最大值和最小值。
% 接下来，我们采用KMeans算法，将不同的通道的进行聚类，得到$g$个cluster。
% 聚类的标准是通道最大值与最小值组成的点。
% 这样，具有相近最大值与最小值的通道将会聚到一起。它们共享量化参数。

\subsection{Avoid Explicit Reordering and Misalignment}

\begin{figure}[tb] % h:here 当前位置 % b bottom % t top % p 浮动
    \centering
    \includegraphics[width=0.96\textwidth]{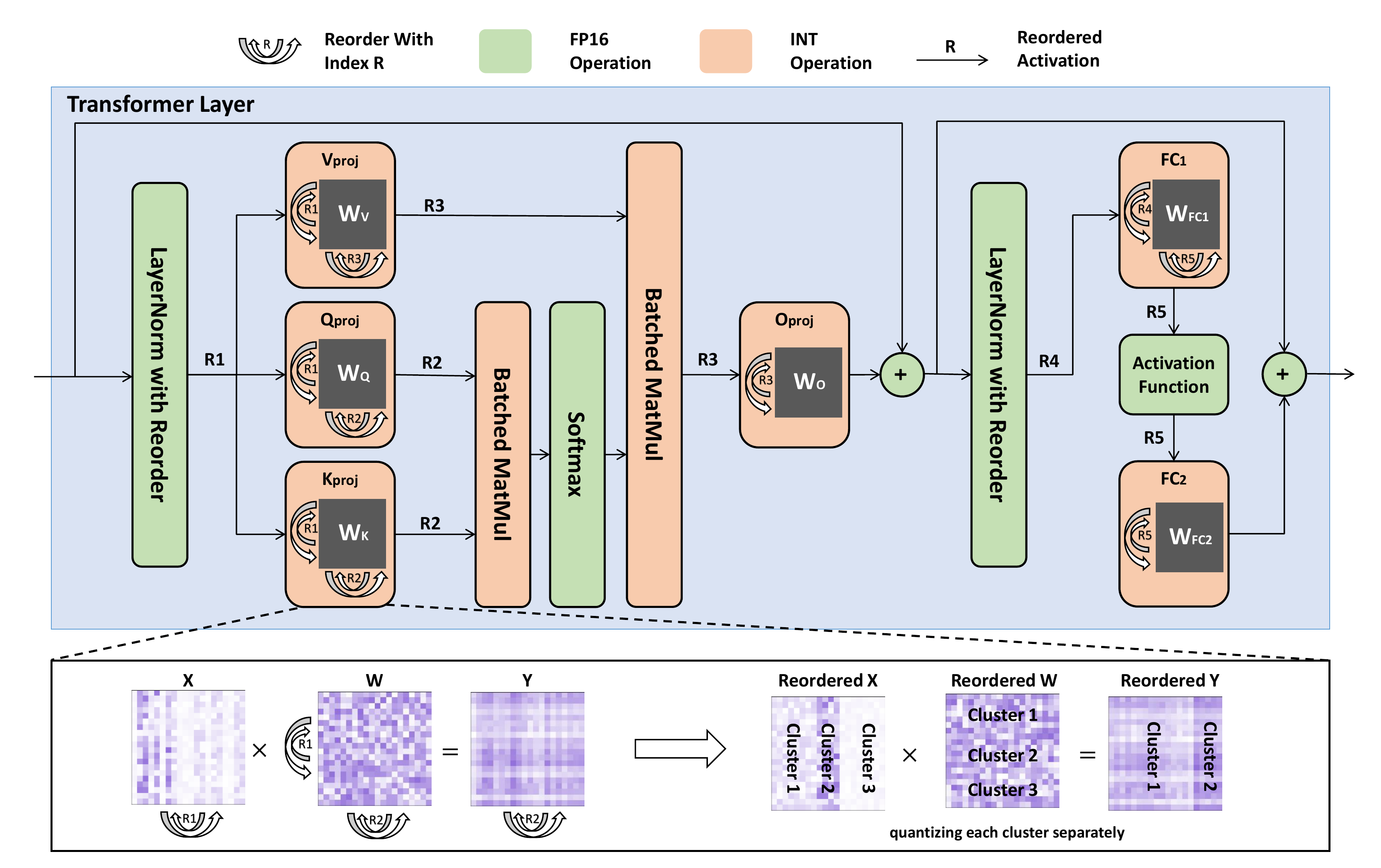} %ims/xx.png
    \caption{An overview of the inference process for a quantized transformer layer with reordered weight and activation. The reordering indexes are represented by the symbols R1 to R5.
    Below the figure illustrates how the weights and activations are reordered in a linear layer, where points with darker colors represent larger values.
    }
    \label{im_overview}
\end{figure}

Explicit reordering is an operation in which the channels in activations are rearranged during run-time by physically relocating the data of different channels from one memory location to another. 
The reordering process will increase inference latency, particularly for large-scale models with a high number of channels. 
Additionally, storing both source and target activation tensors contributes to memory overhead. 
To minimize the overhead of the reorder operation, we fuse it into other operations.

Firstly, we fuse the reorder operation into the layer norm operation. 
Specifically, after computing the layer norm results for each channel, the results are written back to DRAM. The write address is additionally offset based on the reorder index.
This enables layer norm to directly store the results as reordered activations without affecting the weight and bias parameters of the transformer layer. 
As illustrated in Figure~\ref{im_overview}, we modify the two layer norm operations in the transformer layer.

% Secondly, we modify the weight parameters of the network to enable linear layers to directly accept reordered activations and output reordered activations. 
% Specifically, let $W\in \mathbb{R}^{C2\times C1}$ be the weight matrix, we adjust the dimension $C1$ to be the reordered indices of the input activation, and the dimension $C2$ to be the reordered indices of the output activation. 
% The original computation of a linear layer can be represented as
% $Y=\text{bmm}(W,X)$.
% where $\text{bmm}$ is batched matrix multiplication.
% After the weight transformation, the new weight matrix $\tilde{W}$ is obtained by rearranging the rows and columns of the original weight matrix $W$ based on the index of the reordered activations. 
% The modified linear layer can be represented as $\tilde{Y}=\text{bmm}(\tilde{W},\tilde{X})$.
% where $\tilde{Y}$ is the output activation.
% The channel ordering of the output tensor $\tilde{Y}$ follows the same order as the channel ordering of dimension $C2$ of the weight.
% This approach reduces the computational overhead associated with explicit reordering and improves the efficiency of the network.
Secondly, we adjust the weight parameters of the network to allow linear layers to directly accept reordered activations and output reordered activations. Let $W\in \mathbb{R}^{C2\times C1}$ be the weight matrix.
The original computation of a linear layer can be expressed as $Y=\text{bmm}(X,W)$, where $\text{bmm}$ denotes batched matrix multiplication. 
We reorder the columns of the weight matrix (dimension $C1$) based on the input activation reorder index, and the rows (dimension $C2$) based on the output activation reorder index.
The new weight matrix $\tilde{W}\in \mathbb{R}^{C2\times C1}$ is obtained by rearranging the rows and columns of the original weight matrix $W$. 
The modified linear layer can be expressed as $\tilde{Y}=\text{bmm}(\tilde{X},\tilde{W})$, where $\tilde{X}$ is the reordered input. 
The channel ordering of output tensor $\tilde{Y}$ adheres to the same order as the channel ordering of dimension $C2$ of the weight. 
Note that the weight reordering can be completed before deployment, resulting in zero overhead related to reordering during inference.

% Misalignment between two tensors' channels refers to a situation where the ordering of channels in one tensor is different from the ordering of channels in the other tensor. This can occur, for example, when applying operations such as matrix multiplication or element-wise addition between tensors with different channel orders. In such cases, misalignment can cause errors in the calculation, leading to incorrect results. It is important to ensure that the channel orders of the tensors are aligned to prevent such errors.
Misalignment between two tensors' channels refers to a situation in which the ordering of channels in one tensor differs from that in the other tensor. This can occur when applying operations such as matrix multiplication or element-wise addition between tensors with different channel orders. In such instances, misalignment can cause errors in calculation, leading to incorrect results. It is crucial to ensure that the tensors' channel orders are aligned to prevent such errors.

% In the Transformer layer, there is a residual connection where the layer input is added to the output of the out projection layer, and the result is added to the output of the final linear layer. 
% If the output of the out projection layer is reordered, then the channels would not align with those of the input. 
% The same holds true for the output of the final linear layer. 
% We do not reorder the output of the out projection layer and the final linear layer to maintain channel consistency with the original input.
In the Transformer layer, there is a residual connection wherein the layer input is added to the output of the out projection layer, and the result is added to the output of the final linear layer. If the out projection layer's output is reordered, the channels would not align with those of the input. The same applies to the output of the final linear layer. 
Therefore, we do not reorder the output of the out projection layer and the final linear layer to maintain channel consistency with the original input.
To maintain channel consistency with the original input, we don't reorder the output of the out projection layer and the final linear layer.

% Lastly, the query $X^Q$ and the key $X^{K}$ in self-attention should share the same reordering index to avoid misalignment in their matrix-multiplied. 
% We compute the maximum and minimum values of each channel in Q and K and combine them into a quaternion point $(X^Q_{\max,i}, X^Q_{\min,i}, X^K_{\max,i}, X^K_{\min,i})$ for K-means clustering. The clustering result is then utilized for reordering in both $X^Q$ and $X^K$.
Lastly, we note that the channel dimensions of query activation $X^Q$ and key activation $X^K$ are multiplied and summed together in the QK-MatMul of self-attention: $X^Q(X^K)^T$.
To avoid misalignment, $X^Q$ and $X^K$ should share the same reorder index. 
We collect the maximum and minimum values of each channel in Q and K and combine them into a quaternion point $(X^Q_{\max,i}, X^Q_{\min,i}, X^K_{\max,i}, X^K_{\min,i})$ for K-means clustering. 
The clustering result is then employed for reordering in both $X^Q$ and $X^K$.
The reordering for each activation are demonstrated in Figure~\ref{im_overview}.

% 我们将网络中的权重参数进行调整，能让linear层能直接接收reordered activations和直接输出reordered activations.
% 具体来说，假设权重是$W\in R^{C1\times C2}$，我们对C1通道调整为输入的activation的reorder的index，对C2通道调整为输出的activation的index。

% Transformer layer中还有residual connection，layer的输入要与out projection层的输出相加。这个结果再次和最后一层linear层输出的结果相加。
% 由于这个加法是element wise的操作，进行通道reordering会造成不正确的结果。
% 因此，我们不对out projection层的输出和最后一层linear层的输出做reorder，以保持通道与原始输入一致。

% There are two tensors that need to share the same reordering index in transformer layer. 
% The query $X^Q\in\mathbb{R}^{B,N,C}$ and transposed key $X^{K^T}\in\mathbb{R}^{B,C,N}$ is matrix-multiplied in batch, and the channel dimension of the transposed key should correspond to the channel dimension of query at the corresponding position. 
% Therefore, it is necessary to ensure that the reorder index of Q and K is the same to avoid misalignment.

% This approach avoids explicit reordering before multiplication of Q and transposed K.

% Finally, transformer layer中还有两个tensor需要share相同的index的情况。例如，Q与K的转置做矩阵乘法，K转置后的通道维度与Q的通道维度对应位置相乘。因此需要保证Q和K的reorder index相同才能正确计算。
% 具体来说，我们分别计算Q和K每个通道的最大值和最小值。
% 然后将它们组成一个四维点之后，使用Kmeans进行clustering。
% 这样，我们能避免Q与K的explicit reorder。

% \subsection{Framework}

\section{Experiments}

\subsection{Settings}

We will evaluate our proposed reorder-based post-training quantization (RPTQ) on OPT models~\cite{zhang2022opt}.
As our work focus on processing the problem in quantizing activations, we use GPTQ~\cite{frantar2022gptq} to quantize the weights in LLMs~\footnote{The GPTQ method is solely designed for weight quantization and cannot be used for activation quantization. For details on how to combine the GPTQ method for weight quantization with the RPTQ method for activation quantization, please refer to the \ifthenelse{\boolean{review_flag}}{supplementary material}{Appendix~\ref{appendix_combine_gptq}}.}.
We apply static quantization to all the weights and input activations. 
For each cluster of activation, we use the Min-Max method to set the parameters of asymmetric quantization (scaling factor $s$ and zero point $z$).
The quantization of the output activation of layer norm and softmax is performed at 8-bit precision. %, and will be discussed in the next section.
We calculate the R2 and R3 indices for each head in self-attention.
We use 256 samples randomly selected from WikiText2~\cite{WikiText2}, Pen Treebank~\cite{PenTreebank}, and C4~\cite{C4} for calibration dataset.
We will report on perplexity, performance on zero-shot tasks, and memory consumption.
Additionally, we conducted an ablation study on the number of clusters to explore the impact on the quantization performance.
The experiments are conducted on a server equipped with 8 Nvidia A6000 GPUs.

% All experiments were conducted on a GPU server with the Nvidia Tesla A100.

\subsection{Results on LLM}

\begin{table}[tb]
\centering
\caption{Perplexity scores of various models under diverse quantization configurations on three datasets: WikiText2 (WIKI), Pen Treebank (PT), and C4.}
\label{tab_ppl}
\resizebox{\linewidth}{!}{
\begin{tabular}{@{}ccccccccccccccccccc@{}}
\toprule
Model  & \multicolumn{3}{c}{OPT-1.3b} & \multicolumn{3}{c}{OPT-6.7b} & \multicolumn{3}{c}{OPT-13b} & \multicolumn{3}{c}{OPT-30b} & \multicolumn{3}{c}{OPT-66b} & \multicolumn{3}{c}{OPT-175b} \\ \midrule
Task   & WIKI     & PT      & C4      & WIKI     & PT      & C4      & WIKI    & PT      & C4      & WIKI    & PT      & C4      & WIKI    & PT      & C4      & WIKI     & PT      & C4      \\ \midrule
FP16   & 14.63    & 16.96   & 14.72   & 10.86    & 13.09   & 11.74   & 10.13   & 12.34   & 11.20   & 9.56    & 11.84   & 10.69   & 9.34    & 11.36   & 10.28   & 8.34     & 12.01   & 10.13   \\
W4A16  & 14.78    & 17.21   & 14.92   & 11.18    & 13.62   & 12.07   & 10.29   & 12.45   & 11.27   & 9.55    & 11.91   & 10.74   & 9.30    & 11.42   & 10.31   & 8.37     & 12.31   & 10.26   \\
W4A8   & 15.39    & 17.79   & 15.48   & 11.21    & 13.74   & 12.11   & 10.90   & 13.40   & 11.62   & 10.22   & 12.41   & 11.01   & 9.46    & 11.73   & 10.57   & 8.43     & 12.24   & 10.49   \\
W4A4   & 16.88    & 19.23   & 16.55   & 12.00    & 15.17   & 12.85   & 12.74   & 15.76   & 14.71   & 11.15   & 14.11   & 13.48   & 12.23   & 18.87   & 15.93   & 10.60    & 15.59   & 12.28   \\
W4A4KV & 15.26    & 17.65   & 15.37   & 11.26    & 13.44   & 12.03   & 10.59   & 12.80   & 11.54   & 9.99    & 12.18   & 11.01   & 9.75    & 11.64   & 10.61   & 8.40     & 12.38   & 10.54   \\
W4A3KV & 17.22    & 19.94   & 16.92   & 11.92    & 14.13   & 12.61   & 11.15   & 13.90   & 12.04   & 11.62   & 14.95   & 11.96   & 10.88   & 14.69   & 11.36   & 9.39     & 13.45   & 11.27   \\
W3A3KV & 18.45    & 21.33   & 18.26   & 12.42    & 14.48   & 13.13   & 11.47   & 14.08   & 12.41   & 11.76   & 14.98   & 12.22   & 11.47   & 15.03   & 11.75   & 10.03    & 13.82   & 11.30   \\ \bottomrule
\end{tabular}
}
\end{table}
\begin{table}[tb]
\centering
\caption{Accuracy of OPT models under diverse quantization configurations on different zero-shot tasks: LAMBADA(OpenAI), PIQA, ARC(Easy), ARC(Challenge),  OpenBookQA, BoolQ.}
\label{tab_acc}
\resizebox{0.82\linewidth}{!}{
\begin{tabular}{@{}ccccccccccc@{}}
\toprule
Task   & \multicolumn{5}{c}{LAMBADA(OpenAI)~\cite{lambada}}  & \multicolumn{5}{c}{PIQA~\cite{piqa}}                     \\ \midrule
Model  & 1.3b     & 6.7b   & 13b    & 30b    & 66b    & 1.3b     & 6.7b     & 13b     & 30b     & 66b     \\ \midrule
FP16   & 57.98\%   & 61.84\% & 68.60\% & 71.41\% & 67.14\% & 72.47\%   & 74.53\%   & 76.87\%  & 78.01\%  & 78.12\%  \\
W4A16  & 57.46\%   & 60.78\% & 68.50\% & 71.37\% & 67.06\% & 71.59\%   & 74.80\%   & 76.93\%  & 78.29\%  & 78.18\%  \\
W4A8   & 52.39\%   & 67.35\% & 62.44\% & 64.99\% & 67.02\% & 69.69\%   & 75.89\%   & 75.46\%  & 76.93\%  & 77.52\%  \\
W4A4   & 49.34\%   & 64.93\% & 60.23\% & 63.92\% & 68.50\% & 68.66\%   & 75.40\%   & 73.55\%  & 76.16\%  & 77.14\%  \\
W4A4KV & 52.90\%   & 67.39\% & 62.77\% & 64.89\% & 69.99\% & 69.26\%   & 76.00\%   & 74.42\%  & 76.65\%  & 76.98\%  \\
W4A3KV & 47.02\%   & 64.97\% & 61.05\% & 59.20\% & 66.23\% & 68.22\%   & 75.73\%   & 73.23\%  & 67.46\%  & 74.21\%  \\
W3A3KV & 42.84\%   & 64.11\% & 60.02\% & 58.33\% & 65.28\% & 68.22\%   & 74.64\%   & 74.10\%  & 67.51\%  & 75.13\%  \\ \midrule
Task   & \multicolumn{5}{c}{ARC(Easy)~\cite{clark2018think}} & \multicolumn{5}{c}{ARC(Challenge)~\cite{clark2018think}} \\ \midrule
Model  & 1.3b     & 6.7b   & 13b    & 30b    & 66b    & 1.3b     & 6.7b     & 13b     & 30b     & 66b     \\ \midrule
FP16   & 51.05\%   & 58.03\% & 61.91\% & 65.31\% & 64.68\% & 29.69\%   & 33.61\%   & 35.66\%  & 38.05\%  & 38.99\%  \\
W4A16  & 51.17\%   & 57.02\% & 61.82\% & 65.10\% & 64.89\% & 30.03\%   & 32.59\%   & 35.49\%  & 37.96\%  & 38.99\%  \\
W4A8   & 48.35\%   & 60.18\% & 60.94\% & 63.46\% & 64.60\% & 26.36\%   & 34.04\%   & 35.58\%  & 37.45\%  & 38.82\%  \\
W4A4   & 47.55\%   & 56.90\% & 58.41\% & 62.12\% & 63.76\% & 25.85\%   & 34.30\%   & 33.95\%  & 36.17\%  & 37.20\%  \\
W4A4KV & 47.76\%   & 57.74\% & 58.54\% & 63.59\% & 63.67\% & 27.64\%   & 33.95\%   & 34.21\%  & 37.37\%  & 37.71\%  \\
W4A3KV & 46.29\%   & 56.69\% & 56.10\% & 48.44\% & 59.00\% & 26.02\%   & 33.95\%   & 33.95\%  & 30.71\%  & 36.77\%  \\
W3A3KV & 44.02\%   & 55.59\% & 53.74\% & 50.42\% & 57.65\% & 26.53\%   & 32.16\%   & 32.50\%  & 30.71\%  & 34.98\%  \\ \midrule
Task   & \multicolumn{5}{c}{OpenBookQA~\cite{openbookqa}}    & \multicolumn{5}{c}{BoolQ~\cite{boolq}}                   \\ \midrule
Model  & 1.3b     & 6.7b   & 13b    & 30b    & 66b    & 1.3b     & 6.7b     & 13b     & 30b     & 66b     \\ \midrule
FP16   & 33.00\%   & 38.00\% & 39.00\% & 40.20\% & 41.60\% & 57.73\%   & 67.03\%   & 65.90\%  & 70.45\%  & 70.85\%  \\
W4A16  & 31.80\%   & 37.40\% & 39.20\% & 40.60\% & 42.00\% & 58.99\%   & 59.72\%   & 66.66\%  & 70.70\%  & 70.55\%  \\
W4A8   & 32.40\%   & 38.00\% & 38.60\% & 39.40\% & 41.80\% & 46.88\%   & 65.93\%   & 66.57\%  & 70.64\%  & 71.07\%  \\
W4A4   & 32.60\%   & 38.40\% & 38.00\% & 38.60\% & 42.00\% & 41.37\%   & 65.44\%   & 58.47\%  & 67.70\%  & 70.24\%  \\
W4A4KV & 32.60\%   & 38.40\% & 38.00\% & 39.80\% & 41.60\% & 43.33\%   & 62.11\%   & 62.47\%  & 68.22\%  & 70.79\%  \\
W4A3KV & 32.80\%   & 36.80\% & 37.00\% & 34.00\% & 39.40\% & 42.84\%   & 61.31\%   & 57.76\%  & 61.74\%  & 67.06\%  \\
W3A3KV & 28.40\%   & 35.20\% & 37.20\% & 32.40\% & 38.60\% & 46.23\%   & 60.79\%   & 65.07\%  & 63.08\%  & 67.49\%  \\ \bottomrule
\end{tabular}
}
\end{table}

We conducted an experiment to evaluate OPT across various model scales. 
Specifically, we evaluated OPT's performance under three distinct bit-width configurations: W4A16, W4A8, and W4A4. 
Here, W4A4 refers to weight quantization with 4 bits and activation quantization with 4 bits. 
Additionally, we developed a new quantization scheme, W4A4KV, W4A3KV, and W3A3KV, focusing solely on quantizing the key cache and value cache, which are the major memory consumers when using large sequence length or batch size.

The same as GPTQ~\cite{frantar2022gptq}, we evaluate the perplexity and the prediction accuracy on various zero shot tasks.
The results are presented in Table~\ref{tab_ppl} and Table~\ref{tab_acc}, respectively.
From the table, we can make the following observations:
In general, the performance of the models tends to decrease as the bit-width for activation quantization decreases. 
For instance, by quantizing OPT-175b, W4A8 achieves a perplexity loss of less than 0.5 and W4A4 achieves a perplexity loss of less than 3.
For the key cache and value cache quantization schemes (W4A4KV, W4A3KV, and W3A3KV), it is noticeable that their performance are better. 
In most cases, the performance of the quantized models is close to the FP16 baseline. 
For instance, by quantizing OPT-175b, W4A4KV achieves a perplexity loss of less than 0.5 and W3A3KV achieves a perplexity loss of less than 1.5.
This suggests that focusing on quantizing key and value caches can be beneficial to maintain the performance while reducing memory consumption.

Other methods, such as SmoothQuant~\cite{xiao2022smoothquant} and PEG~\cite{bondarenko2021understanding}, encounters difficulties when attempting to push quantization to 4 bits. {See \ifthenelse{\boolean{review_flag}}{supplementary material}{Appendix~\ref{appendix_comparing}} for detail}. 
The ignorance of range difference prevent them from successfully quantizing activations of LLMs at low bit-widths.
By carefully considering the range distribution in activation values, our method achieves a significant breakthrough in quantizing LLMs with 3-bit activation quantization.

\subsection{Memory Consumption}

\begin{table}[tb]
\centering
\caption{Memory consumption (GB) of LLMs on different batch sizes and sequence lengths.}
\label{tab_memory}
\resizebox{0.8\linewidth}{!}{
\begin{tabular}{@{}ccccccccccc@{}}
\toprule
                          & Batch Size      & \multicolumn{3}{c}{1} & \multicolumn{3}{c}{8} & \multicolumn{3}{c}{64}  \\ \midrule
                          & Sequence Length & 2048  & 4096  & 8192  & 2048  & 4096  & 8192  & 2048  & 4096   & 8192   \\ \midrule
\multirow{7}{*}{OPT-30b}  & W16A16          & 59.4  & 62.3  & 68.1  & 79.7  & 102.9 & 149.3 & 242.0 & 427.5  & 798.6  \\
                          & W4A16           & 17.0  & 19.9  & 25.7  & 37.3  & 60.5  & 106.9 & 199.6 & 385.2  & 756.2  \\
                          & W4A8            & 15.6  & 17.1  & 20.1  & 26.0  & 38.0  & 61.8  & 109.5 & 204.9  & 395.7  \\
                          & W4A4            & 14.9  & 15.7  & 17.3  & 20.4  & 26.7  & 39.3  & 64.5  & 114.8  & 215.4  \\
                          & W4A4KV          & 15.0  & 15.9  & 17.7  & 21.2  & 28.3  & 42.6  & 71.0  & 127.9  & 241.7  \\
                          & W4A3KV          & 14.8  & 15.6  & 17.0  & 19.9  & 25.7  & 37.2  & 60.3  & 106.5  & 198.8  \\
                          & W3A3KV          & 11.3  & 12.0  & 13.5  & 16.4  & 22.1  & 33.7  & 56.8  & 102.9  & 195.3  \\ \midrule
\multirow{7}{*}{OPT-66b}  & W16A16          & 128.1 & 133.0 & 142.7 & 162.1 & 200.9 & 278.5 & 433.8 & 744.3  & 1365.3 \\
                          & W4A16           & 35.7  & 40.5  & 50.2  & 69.6  & 108.4 & 186.1 & 341.3 & 651.9  & 1272.9 \\
                          & W4A8            & 33.3  & 35.8  & 40.7  & 50.6  & 70.5  & 110.1 & 189.5 & 348.1  & 665.4  \\
                          & W4A4            & 32.1  & 33.4  & 36.0  & 41.2  & 51.5  & 72.2  & 113.5 & 196.2  & 361.6  \\
                          & W4A4KV          & 32.2  & 33.7  & 36.5  & 42.2  & 53.6  & 76.4  & 122.0 & 213.1  & 395.4  \\
                          & W4A3KV          & 32.0  & 33.1  & 35.4  & 39.9  & 49.0  & 67.2  & 103.7 & 176.5  & 322.3  \\
                          & W3A3KV          & 24.3  & 25.4  & 27.7  & 32.2  & 41.3  & 59.5  & 96.0  & 168.8  & 314.6  \\ \midrule
\multirow{7}{*}{OPT-175b} & W16A16          & 335.4 & 344.9 & 363.8 & 401.7 & 477.5 & 629.0 & 932.0 & 1538.0 & 2750.1 \\
                          & W4A16           & 91.0  & 100.4 & 119.4 & 157.2 & 233.0 & 384.5 & 687.5 & 1293.5 & 2505.6 \\
                          & W4A8            & 86.3  & 91.1  & 100.7 & 119.9 & 158.4 & 235.3 & 389.0 & 696.5  & 1311.6 \\
                          & W4A4            & 84.0  & 86.4  & 91.4  & 101.3 & 121.1 & 160.6 & 239.8 & 398.0  & 714.6  \\
                          & W4A4KV          & 84.1  & 86.8  & 92.1  & 102.7 & 123.9 & 166.3 & 251.0 & 420.5  & 759.6  \\
                          & W4A3KV          & 83.6  & 85.7  & 89.8  & 98.1  & 114.8 & 148.1 & 214.6 & 347.8  & 614.1  \\
                          & W3A3KV          & 63.2  & 65.3  & 69.4  & 77.8  & 94.4  & 127.7 & 194.3 & 327.4  & 593.7  \\ \bottomrule
\end{tabular}
}
\end{table}

The huge memory consumption is a major challenge in the deployment of LLMs.
Limited memory capacity can result in significant performance bottlenecks~\cite{sheng2023high}.
There are three sources contributing to the memory usage of LLMs: Firstly, the weights in LLMs should be saved in memory, which can be significantly reduced through weight quantization. 
Secondly, memory is required for temporary activations generated during network execution.
As these temporary activations can be released after usage and the memory usage of attention matrices can be greatly reduced through operation fusion~\cite{dao2022flashattention}, their memory footprint is minimal.
Lastly, caching of key and value activations is necessary for predicting subsequent words. 
It is noteworthy that the key and value caches consume a majority of the memory when batch size and sequence length are high. {See \ifthenelse{\boolean{review_flag}}{supplementary material}{Appendix~\ref{appendix_mem_ratio}} for details.}

Table~\ref{tab_memory} presents the memory usage under various settings, where we observe that lower-bit activations can substantially reduce the memory usage, particularly when the batch size and sequence length are high. 
For instance, we observe that W4A8 can reduce about 63\% memory and W4A4 can reduce about 75\% memory.
Therefore, adopting activation quantization can greatly reduce the memory pressure in long-text tasks or large-batch scenarios.
Quantizing solely the key and value cache also considerably diminishes memory consumption. 
We observe that W4A4KV can reduce about 73\% memory and W3A3KV can reduce about 80\% memory.

% 存储开销问题是LLM部署的一大难题，当存储不够的时候，将会造成很大的性能瓶颈。
% 首先，LLM中的权重需要存储在设备内存中。对权重的量化可以很大程度减少存储的使用。
% 其次，网络运行过程中动态产生的activation需要存储在设备内存中。
% 最后，生成式任务中，网络的key activation和value activation需要被cache住，以备下一个单词的预测使用。
% 影响activation存储使用量的因素有batch size和sequence length。
% Figure~\ref{tab_memory} 显示了在不同量化参数下的网络内存使用量。当batch size和sequence length较高是，更低比特的activation量化的对于减少内存使用有很大的帮助。
% 例如，当batch size=16，sequence length=4096时，OPT-66b下W4A4相比于W4A8节省了40GB的存储。

\subsection{Ablation Study}

\begin{figure}[tb] % h:here 当前位置 % b bottom % t top % p 浮动
    \centering
    \includegraphics[width=\textwidth]{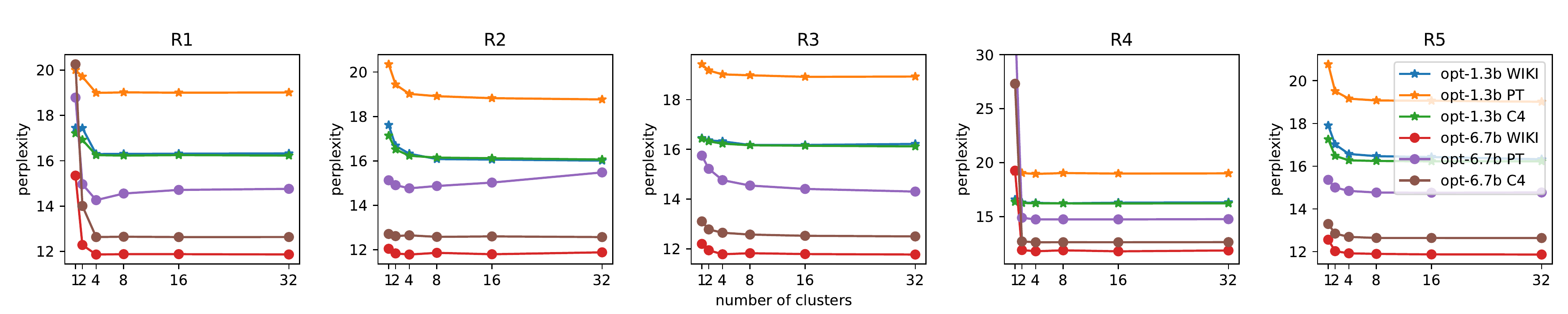} %ims/xx.png
    \caption{The ablation study to evaluate the performance of the clustering method under the W16A4 configuration. We tested different numbers of clusters (1, 2, 4, 8, and 32) for R1 to R5.
    % See appendix for more results.
    }
    \vspace{-10pt}
    \label{im_ablation}
\end{figure}

In this study, we perform an ablation analysis to investigate the impact of varying the number of clusters on model performance. Figure~\ref{im_ablation} presents the results of adjusting the number of clusters for each reorder (R1 to R5) while keeping other reorders fixed.
As the number of clusters increases, the perplexity generally decreases. 
% We noticed that the results showed some fluctuations. In the R2 experiment, the best performance was obtained with four clusters. In the R3 experiment, the model exhibited improved performance with a larger number of clusters. We found that the turning point was when the number of clusters was equal to four. In the R4 experiment, if the number of clusters was set to one, there would be a significant performance loss. As the number of clusters in R4 and R5 increased, there was a slight improvement in performance.
The fluctuations observed in R2 are an intriguing problem. 
We have found that increasing the size of the calibration dataset is helpful in alleviating this issue. {See \ifthenelse{\boolean{review_flag}}{supplementary material}{Appendix~\ref{appendix_ablation}} for details.}.  We suspect that this could be due to the limited amount of calibration data, which may not accurately capture the data distribution of certain samples. 
We have also noticed that larger networks are more sensitive to the number of clusters. 
For instance, if the number of clusters is set to 1, larger networks may crash, while smaller networks may have a lower quantization error.
Because RPTQ reorder each self-attention heads separately, the overhead associated with reordering for each head in R2 and R3 is substantial when the number of self-attention heads is high.
In our experiments, we utilize 32 clusters for R1, R4, and R5, and 4 clusters for R2 and R3.

% We conduct ablation study on number of clusters. 
% As shown in Figure~\ref{im_ablation}, we adjust the number of clusters in each reorder.
% We fix the other reorders and only change the number of clusters from R1 to R5.
% As the number of clusters increase, the perplexity decreases in most cases.
% For R1, we found number of clusters equal to 8 is enough.
% For R2 we found number of clusters equal to 4 is the best.
% For R3, larger is better.
% For R4, 8 is enough.
% For R5, 32 is the best.
% Because we should reorder for each head in R2 and R3, the overhead is high when number of clusters in R2 and R3 is high.
% In our experiment, we use 32 for R1, R4 and R5, and we use 4 for R2 and R3.
% 只量化R1 ncluster=1 2 4 8...32
% 只量化R2

% Min Max 是不是要用Min 

% 

\section{Conclusion}

In this paper, we identify that the main challenge in quantizing large-scale language models (LLMs) stems from the differences value ranges across channels, rather than just the issue of outliers.
We have proposed a novel reorder-based quantization approach, RPTQ, that involves rearranging the channels in the activations and quantizing them in clusters. 
By quantizing the weights and activations, we have significantly reduced the memory usage. 
Our experiments demonstrate that our proposed approach successfully addresses the issue of numerical differences among activation channels and achieves a significant breakthrough by quantizing LLM to 3 bit activation for the first time.

\bibliographystyle{plain}
\bibliography{ref}
\newpage
\appendix

\section{Appendix}

\subsection{Distribution of different channels}
\label{appendix_distribution}

\begin{figure}[tb] % h:here 当前位置 % b bottom % t top % p 浮动
    \centering
    \includegraphics[width=0.85\textwidth]{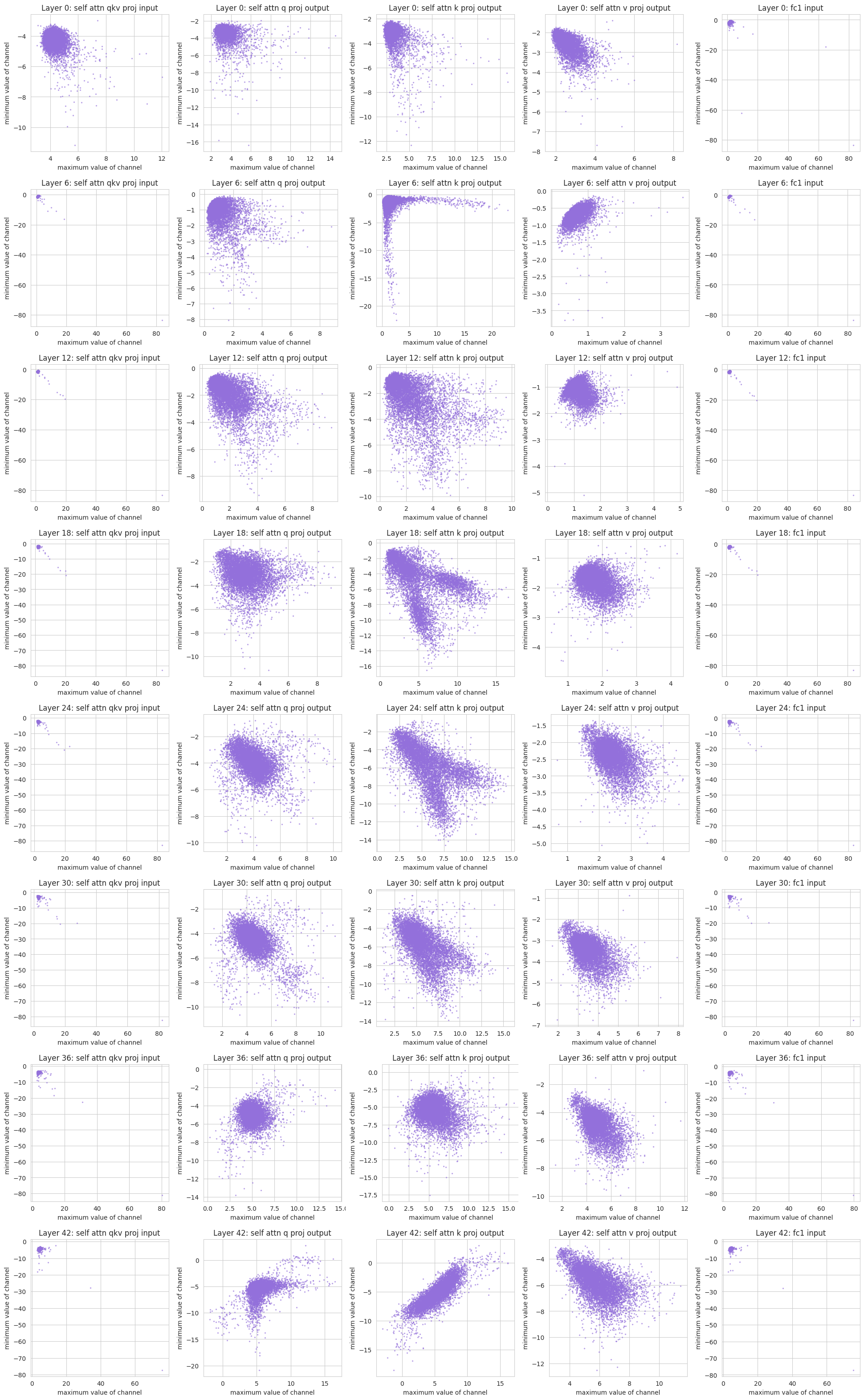} %ims/xx.png
    \caption{Demonstration of the distribution of different channels in OPT-30b decoder layers. Each point is (maximum value, minimum value) of a channel in the activation. 
    }
    \vspace{-10pt}
    \label{im_appendix_distribution_30}
\end{figure}

\begin{figure}[tb] % h:here 当前位置 % b bottom % t top % p 浮动
    \centering
    \includegraphics[width=0.72\textwidth]{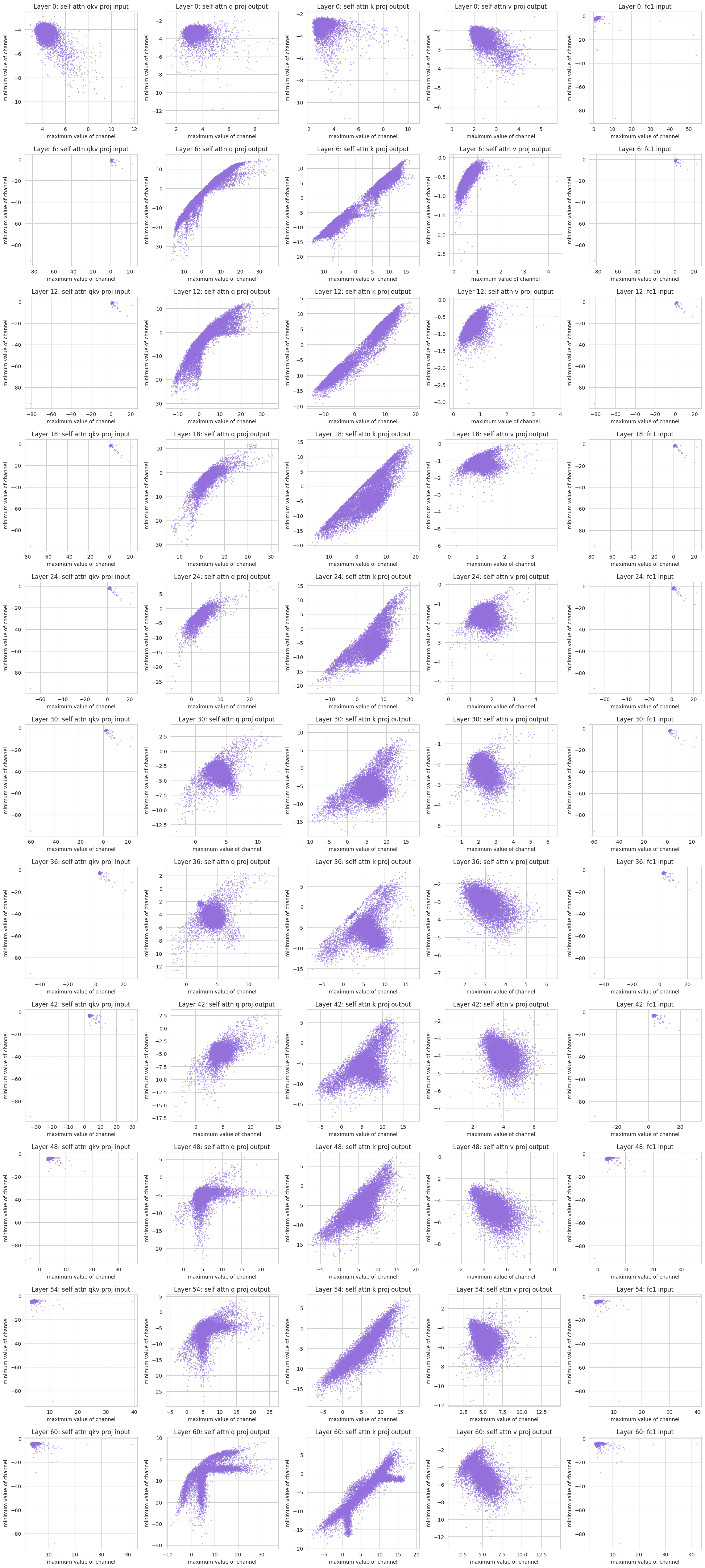} %ims/xx.png
    \caption{Demonstration of the distribution of different channels in OPT-66b decoder layers. Each point is (maximum value, minimum value) of a channel in the activation. 
    }
    \vspace{-10pt}
    \label{im_appendix_distribution_66}
\end{figure}

We analyzed the distribution of different channels in the OPT-30b and OPT-66b by plotting the (maximum value, minimum value) points of each channel in the activation, demonstrated in Figure~\ref{im_appendix_distribution_30} and Figure~\ref{im_appendix_distribution_66} Our analysis revealed that there were significant differences in the data range across different layers and even within different channels in the same layer. This finding motivated us to develop a clustering approach to reduce the impact of these differences during quantization. By clustering channels with similar magnitude ranges together, we can reduce the quantization error and improve the efficiency of quantized LLMs. 

\subsection{Comparing with Other Methods}
\label{appendix_comparing}

\begin{table}[]
\centering
\caption{Comparing RPTQ with SmoothQuant on perplexity scores of various models under diverse quantization configurations on three datasets: WikiText2 (WIKI), Pen Treebank (PT), and C4.}
\label{tab_compare}
\resizebox{0.98\linewidth}{!}{
\begin{tabular}{@{}cccccccccccccc@{}}
\toprule
\multicolumn{2}{c}{Model}             & \multicolumn{3}{c}{OPT-1.3b} & \multicolumn{3}{c}{OPT-6.7b} & \multicolumn{3}{c}{OPT-13b} & \multicolumn{3}{c}{OPT-30b}    \\ \midrule
\multicolumn{2}{c}{Task}              & WIKI     & PT      & C4      & WIKI     & PT      & C4      & WIKI    & PT      & C4      & WIKI     & PT       & C4       \\ \midrule
\multirow{2}{*}{W4A8}   & SmoothQuant & 16.89    & 19.35   & 16.26   & 11.62    & 14.04   & 12.47   & 12.55   & 14.73   & 12.2    & 9.96     & 12.12    & 11.01    \\
                        & RPTQ        & 15.39    & 17.79   & 15.48   & 11.21    & 13.74   & 12.11   & 10.90   & 13.40   & 11.62   & 10.22    & 12.41    & 11.01    \\ \midrule
\multirow{2}{*}{W4A4}   & SmoothQuant & 68.25    & 68.44   & 53.77   & 51.19    & 67.14   & 73.44   & 235.38  & 285.85  & 187.96  & 18435.66 & 70175.91 & 11297.63 \\
                        & RPTQ        & 16.88    & 19.23   & 16.55   & 12.00    & 15.17   & 12.85   & 12.74   & 15.76   & 14.71   & 11.15    & 14.11    & 13.48    \\ \midrule
\multirow{2}{*}{W4A4KV} & SmoothQuant & 19.45    & 22394   & 18.46   & 12.68    & 16.13   & 13.72   & 11.62   & 16.44   & 12.21   & 11.61    & 15.43    & 11.74    \\
                        & RPTQ        & 15.26    & 17.65   & 15.37   & 11.26    & 13.44   & 12.03   & 10.59   & 12.80   & 11.54   & 9.99     & 12.18    & 11.01    \\ \bottomrule
\end{tabular}
}
\end{table}

\begin{table}[]
\centering
\caption{Comparing RPTQ with PEG on perplexity scores of various models under W4A4 on three datasets: WikiText2 (WIKI), Pen Treebank (PT), and C4.}
\label{tab_compare_peg}
\resizebox{0.98\linewidth}{!}{
\begin{tabular}{@{}cccccccccccccccc@{}}
\toprule
Model & \multicolumn{3}{c}{OPT-1.3b} & \multicolumn{3}{c}{OPT-6.7b} & \multicolumn{3}{c}{OPT-13b} & \multicolumn{3}{c}{OPT-30b} & \multicolumn{3}{c}{OPT-66b} \\ \midrule
Task  & WIKI     & PT      & C4      & WIKI     & PT      & C4      & WIKI    & PT      & C4      & WIKI    & PT      & C4      & WIKI    & PT      & C4      \\
PEG   & 19.18    & 22.44   & 18.57   & 12.30    & 15.04   & 13.61   & 14.28   & 17.78   & 19.92   & 17.87   & 34.94   & 45.43   & 15.57   & 23.59   & 20.36   \\
RPTQ  & 16.88    & 19.23   & 16.55   & 12.00    & 15.17   & 12.85   & 12.74   & 15.76   & 14.71   & 11.15   & 14.11   & 13.48   & 12.23   & 18.87   & 15.93   \\ \bottomrule
\end{tabular}
}
\end{table}

% As shown in Figure~\ref{im_comparing}，我们将RPTQ与SmoothQuant进行对比，发现SmoothQuant很明显的在较大网络上有很大的性能损失。这是由于大网络上不同通道之间的range差异会比小网络要大。
As depicted in Table~\ref{tab_compare}, we compared RPTQ with SmoothQuant~\cite{xiao2022smoothquant}.
SmoothQuant tackles the quantization difficulty by introducing a mathematically equivalent per-channel scaling transformation that smooths the magnitude across channels, making the activations more amenable to quantization. 
It cannot addressing the high difference of value ranges across channels.
Additionally, this approach may also lead to issues with weight quantization.
We observed that SmoothQuant exhibits significant performance degradation on larger networks. 
This can be attributed to the larger range differences among different channels in large networks compared to smaller networks.
As depicted in Table~\ref{tab_compare_peg}, we compared RPTQ with PEG~\cite{bondarenko2021understanding}.
Due to the original paper was only tested on small models, we applied its method to the OPT model and used the same group settings for PEG as in RPTQ.
It was observed that PEG also incurs significant quantization loss on large models.
% 由于原始PEG仅在小模型上进行测试，我们将其方法迁移到了OPT模型上，并测试其性能。
% 我们可以看到PEG在大模型上也有较大量化损失。

\subsection{Combine RPTQ with GPTQ}
\label{appendix_combine_gptq}

Generative Pre-trained Transformer Quantization (GPTQ) is a post-training quantization method for Generative Pre-trained Transformers (GPTs) that focuses on quantizing the weights in the learner layers of transformers. The primary goal of GPTQ is to minimize the error introduced by quantizing weights. This is achieved through the optimization target:
\begin{equation}
    \text{argmin}_{\hat{W}}||XW-X\hat{W}||_2^2,
\end{equation}
where $W$ is the weight of a linear layer, and $X$ is the input of the layer collected using a calibration dataset. 
By solving this layer-wise quantization problem with a Hessian-based approximation, GPTQ can find the quantized weight that minimizes the quantization error.

To achieve different quantization for each input cluster of weight, we apply GPTQ on each input cluster of weights in the layer to minimize the quantization error. To combine Random Perturbation Training Quantization (RPTQ) with GPTQ, we first reorder the weights in the linear layers before applying GPTQ.

\subsection{Computation under cluster-based quantization}
\label{appendix_computation}

% 在RPTQ量化网络之后，每个activation $X$调整通道变成$\tilde{X}$，它在通道上被划分成了$g$个clusters，每个cluster的asymmetric 量化参数（scale s和zero point）不同。let 第i个cluster的scale为$s_i^X$，zero point为$z_i^X$。下一层的weight $W$也按照输入activation的index对权重的输入维度进行调整成$\tilde{X}$。也划分成为$g$个clusters，每个cluster的asymmetric 量化参数（scale s和zero point）不同。let 第i个cluster的scale为$s_i^W$，zero point为$z_i^W$。在计算时，不同cluster中的activation和weight进行矩阵乘法，结果反量化后，不同cluster的结果加到一起。
After quantizing the network with RPTQ, each activation $X$ is adjusted by channel to become $\tilde{X}$, which is divided into $g$ clusters.
Each activation cluster $\tilde{X}_i$ has different asymmetric quantization parameters (scale and zero point). 
We denote the scale and zero point for the $i$-th cluster as $s_i^X$ and $z_i^X$, respectively. 
The activations in the $i$-th cluster are quantized with the quantization parameters:
\begin{equation}
    \tilde{X}_{q,i}=Q_k(\tilde{X}_i,s_i^X, z_i^X),
\end{equation}
where $Q_k$ represents the quantization function that quantizes the input using the $k$-bit quantization parameters. 
The weights $W$ in the next layer are also adjusted based on the input activation's reorder index to become $\tilde{W}$. It is also divided into $g$ clusters, where each cluster has different asymmetric quantization parameters. 
For one output channel, each weight cluster $\tilde{W}_i$ is also quantized with different asymmetric quantization parameters (scale and zero point), which are denoted as $s_i^W$ and $z_i^W$, respectively. The weights in the $i$-th cluster are quantized with the quantization parameters:
\begin{equation}
    \tilde{W}_{q,i}=Q_k(\tilde{W}_i,s_i^W, z_i^W).
\end{equation}
This process ensures that both the activations and weights are quantized with different quantization parameters for each cluster.

We describe two methods for computing the output tensor $Y$ using the quantized activations and weights. The first method dequantizes the weight and activation values back to floating-point numbers, while the second method performs computation directly in the integer domain.
The first method involves dequantizing the quantized activations and weights back to floating-point values as follows:
\begin{equation}
\hat{X}i = s^X_i(X{q,i} - z^X_i), \quad \hat{W}i = s^W_i(W{q,i} - z^W_i).
\end{equation}
The dequantized values are then concatenated to form the full activations $\hat{X}$ and the full weights $\hat{W}$. Matrix multiplication is performed using the dequantized values:
\begin{equation}
Y = \hat{X}\hat{W}.
\end{equation}

The second method computes the output tensor $Y$ directly in the integer domain. The quantized activations and weights in each cluster are multiplied using matrix and vector multiplication:
\begin{equation}
Y_{q,i} = \tilde{X}{q,i}\tilde{W}{q,i} - z^X_i\tilde{W}{q,i} - z^W_i\tilde{X}{q,i} + z^X_iz^W_i.
\end{equation}
The results are then dequantized, and the dequantized results are summed up:
\begin{equation}
Y = \sum_{i=1}^g (s^X_i s^W_i Y_{q,i}).
\end{equation}
The second method is computationally efficient, but it requires hardware support for integer arithmetic. For instance, the 3-bit value is not supported in most GPUs.
In such cases, these values are cast to higher hardware-supported datatypes such as 4-bit or 8-bit on GPUs.

% There are two methods to achieve the computation.
% The first one is de-quantize the weight and activation to floating point values:
% \begin{equation}
%     \hat{X}_i=s^X_i(X_{q,i}-z^X_i),\quad \hat{W}_i=s^W_i(W_{q,i}-z^W_i).
% \end{equation}
% Then the de-quantized values are concatenated to the full activations $\hat{X}$ and the full weights $\hat{W}$.
% We perform matrix multiplication with the de-quantized values:
% \begin{equation}
%     Y=\hat{X}\hat{W}.
% \end{equation}
% The second method is computed in integer values.
% The quantized activations and quantized weights in each cluster are multiplied using matrix multiplication and vector multiplication:
% \begin{equation}
%     Y_{q,i}=\tilde{X}_{q,i}\tilde{W}_{q,i}-z^X_i\tilde{W}_{q,i}-z^W_i\tilde{X}_{q,i}+z^X_iz^W_i.
% \end{equation}
% Then the results are being de-quantized and the de-quantized results are summed up:
% \begin{equation}
%     Y=\sum_{i=1}^g (s^X_i s^W_i Y_{q,i}).
% \end{equation}
% The second method is computational efficient. But it requires hardware support. For example, the 3 bit activation is not supported in most GPUs.

\subsection{More Results for Ablation Study}
\label{appendix_ablation}

\begin{figure}[tbp]
	\centering
    \subfloat[OPT-6.7b with 128 calibration samples]{\includegraphics[width=\textwidth]{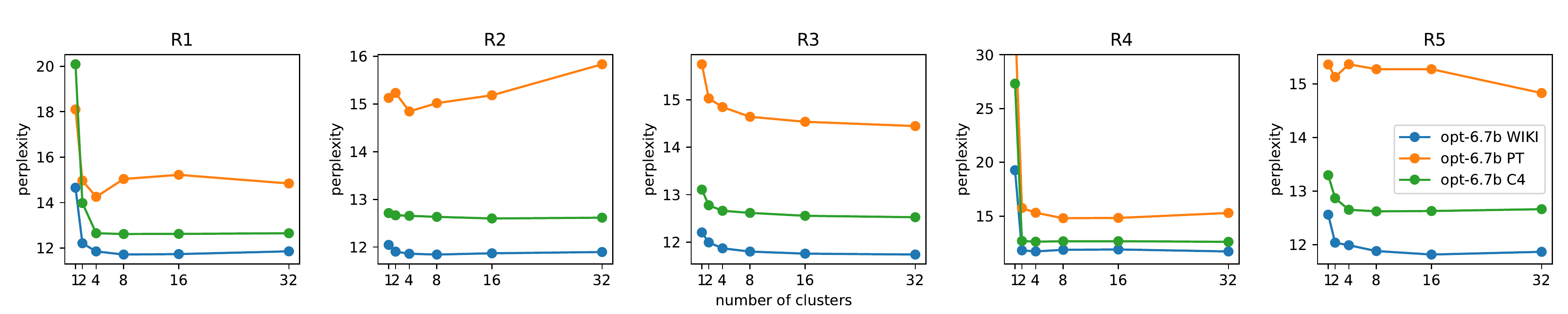}}\\
	\subfloat[OPT-6.7b with 256 calibration samples]{\includegraphics[width=\textwidth]{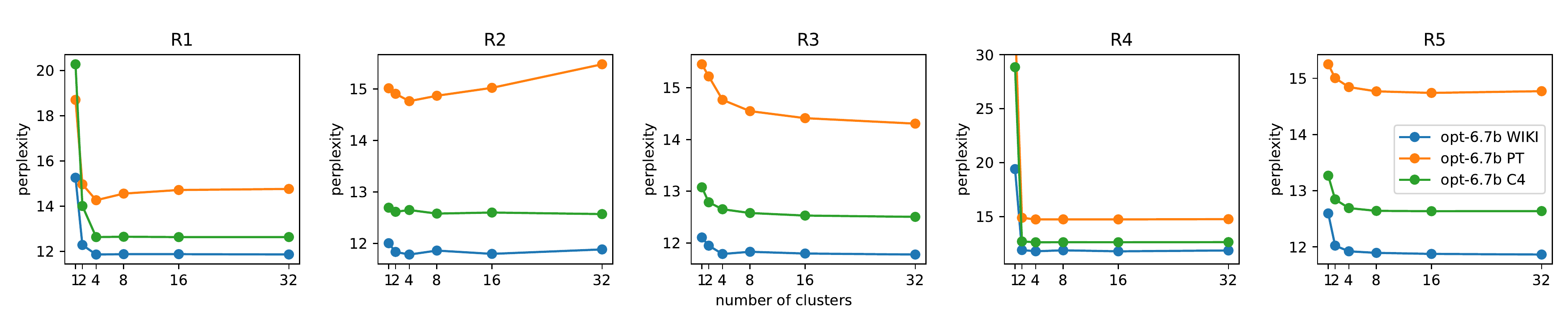}}\\
	\subfloat[OPT-13b with 128 calibration samples]{\includegraphics[width=\textwidth]{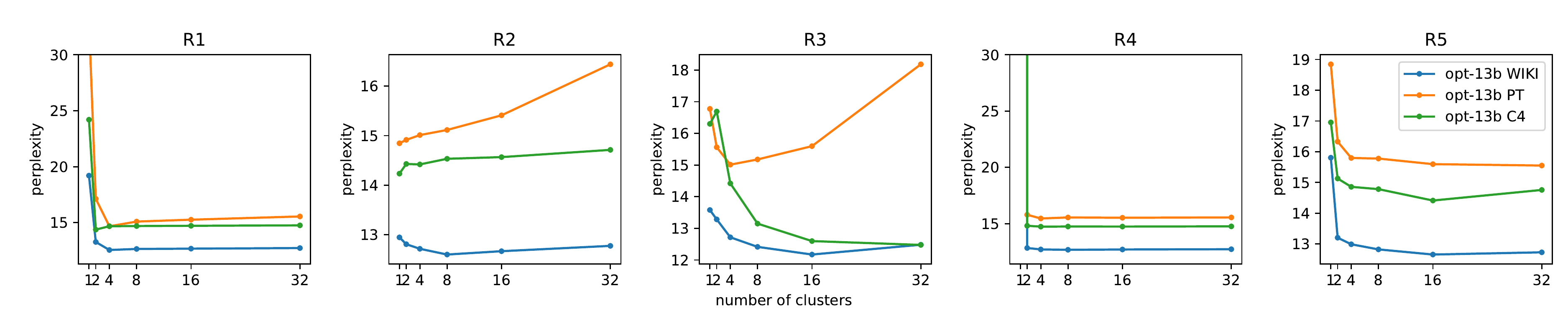}}\\
	\subfloat[OPT-13b with 256 calibration samples]{\includegraphics[width=\textwidth]{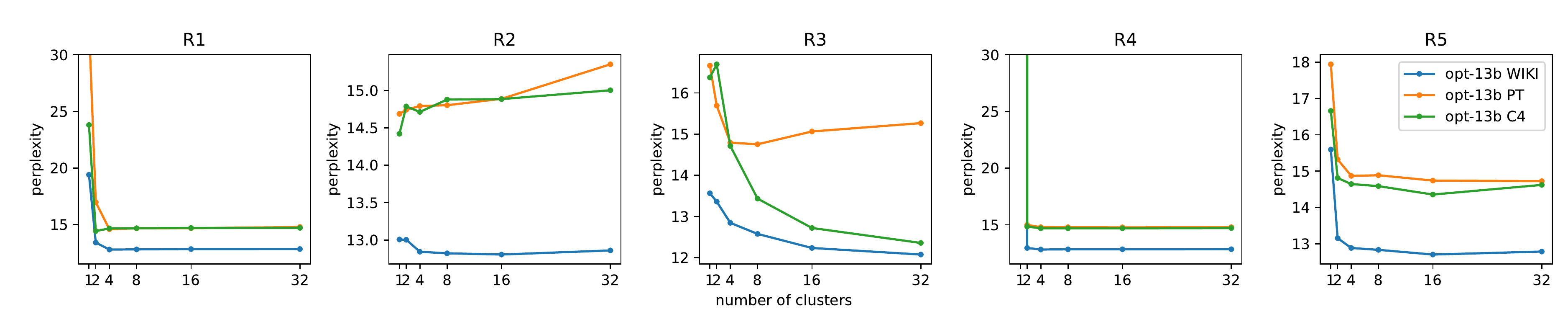}}\\
	\caption{The ablation study to evaluate the performance of the clustering method under the W16A4 configuration. We tested different numbers of clusters (1, 2, 4, 8, and 32) for R1 to R5.}
 \label{im_ablation_nsamples}
\end{figure}

Figure~\ref{im_ablation_nsamples} displays the outcomes associated with 128/256 calibration samples. The observed fluctuations in PT and C4 within R2 and R3 present an intriguing issue. We hypothesize that these fluctuations may be attributable to the limited quantity of calibration data, which might not accurately represent the data distribution for specific samples.
However, using a larger calibration dataset during the calibration phase would demand more memory and computational resources, which is why we have not yet conducted this experiment in this paper. 
In the future, with additional resources, we intend to conduct more extensive analyses using larger networks and calibration datasets to gain a deeper understanding of this matter.

\subsection{Memory Consumption of Different Parts}
\label{appendix_mem_ratio}

\begin{table}[]
\centering
\caption{The memory proportion of different parts in LLMs.}
\label{tab_memory_proportion}
\resizebox{1\linewidth}{!}{
\begin{tabular}{@{}cccccccccccccc@{}}
\toprule
\multicolumn{2}{c}{Batch Size}        & \multicolumn{6}{c}{1}                                                 & \multicolumn{6}{c}{64}                                                \\ \midrule
\multicolumn{2}{c}{Sequence Length}   & \multicolumn{3}{c}{2048}          & \multicolumn{3}{c}{8192}          & \multicolumn{3}{c}{2048}          & \multicolumn{3}{c}{8192}          \\ \midrule
Model                     & Precision & Weight  & K/V & Dynamic & Weight  & K/V & Dynamic & Weight  & K/V & Dynamic & Weight  & K/V & Dynamic \\ \midrule
\multirow{7}{*}{OPT-1.3b} & FP16      & 85.35\% & 12.12\%   & 2.53\%      & 59.30\% & 33.67\%   & 7.03\%      & 8.35\%  & 75.83\%   & 15.82\%     & 2.23\%  & 80.89\%   & 16.88\%     \\
                          & W4A16     & 59.30\% & 33.67\%   & 7.03\%      & 26.70\% & 60.65\%   & 12.65\%     & 2.23\%  & 80.89\%   & 16.88\%     & 0.57\%  & 82.27\%   & 17.17\%     \\
                          & W4A8      & 73.48\% & 20.86\%   & 5.66\%      & 40.92\% & 46.47\%   & 12.62\%     & 4.15\%  & 75.39\%   & 20.47\%     & 1.07\%  & 77.81\%   & 21.12\%     \\
                          & W4A4      & 83.45\% & 11.85\%   & 4.70\%      & 55.76\% & 31.66\%   & 12.57\%     & 7.30\%  & 66.35\%   & 26.35\%     & 1.93\%  & 70.19\%   & 27.88\%     \\
                          & W4A4KV    & 80.47\% & 11.42\%   & 8.11\%      & 50.74\% & 28.81\%   & 20.45\%     & 6.05\%  & 54.95\%   & 39.00\%     & 1.58\%  & 57.57\%   & 40.85\%     \\
                          & W4A3KV    & 82.94\% & 8.83\%    & 8.23\%      & 54.86\% & 23.36\%   & 21.78\%     & 7.06\%  & 48.10\%   & 44.84\%     & 1.86\%  & 50.79\%   & 47.35\%     \\
                          & W3A3KV    & 78.47\% & 11.14\%   & 10.39\%     & 47.68\% & 27.08\%   & 25.24\%     & 5.39\%  & 48.96\%   & 45.65\%     & 1.40\%  & 51.03\%   & 47.57\%     \\ \midrule
\multirow{7}{*}{OPT-6.7b} & FP16      & 91.70\% & 7.17\%    & 1.12\%      & 73.43\% & 22.98\%   & 3.59\%      & 14.73\% & 73.74\%   & 11.53\%     & 4.14\%  & 82.90\%   & 12.96\%     \\
                          & W4A16     & 73.43\% & 22.98\%   & 3.59\%      & 40.86\% & 51.14\%   & 8.00\%      & 4.14\%  & 82.90\%   & 12.96\%     & 1.07\%  & 85.55\%   & 13.38\%     \\
                          & W4A8      & 84.16\% & 13.17\%   & 2.68\%      & 57.04\% & 35.70\%   & 7.26\%      & 7.66\%  & 76.73\%   & 15.60\%     & 2.03\%  & 81.41\%   & 16.56\%     \\
                          & W4A4      & 90.79\% & 7.10\%    & 2.11\%      & 71.13\% & 22.26\%   & 6.62\%      & 13.34\% & 66.80\%   & 19.86\%     & 3.71\%  & 74.22\%   & 22.07\%     \\
                          & W4A4KV    & 89.30\% & 6.99\%    & 3.71\%      & 67.60\% & 21.15\%   & 11.25\%     & 11.54\% & 57.75\%   & 30.71\%     & 3.16\%  & 63.22\%   & 33.62\%     \\
                          & W4A3KV    & 90.94\% & 5.34\%    & 3.73\%      & 71.50\% & 16.78\%   & 11.72\%     & 13.55\% & 50.89\%   & 35.55\%     & 3.77\%  & 56.65\%   & 39.58\%     \\
                          & W3A3KV    & 88.27\% & 6.91\%    & 4.82\%      & 65.30\% & 20.43\%   & 14.27\%     & 10.52\% & 52.68\%   & 36.80\%     & 2.86\%  & 57.19\%   & 39.95\%     \\ \midrule
\multirow{7}{*}{OPT-13b}  & FP16      & 93.28\% & 5.97\%    & 0.75\%      & 77.64\% & 19.87\%   & 2.49\%      & 17.83\% & 73.03\%   & 9.14\%      & 5.15\%  & 84.31\%   & 10.55\%     \\
                          & W4A16     & 77.64\% & 19.87\%   & 2.49\%      & 46.47\% & 47.58\%   & 5.95\%      & 5.15\%  & 84.31\%   & 10.55\%     & 1.34\%  & 87.69\%   & 10.97\%     \\
                          & W4A8      & 87.05\% & 11.14\%   & 1.81\%      & 62.69\% & 32.09\%   & 5.22\%      & 9.50\%  & 77.83\%   & 12.66\%     & 2.56\%  & 83.81\%   & 13.64\%     \\
                          & W4A4      & 92.66\% & 5.93\%    & 1.41\%      & 75.94\% & 19.44\%   & 4.62\%      & 16.47\% & 67.47\%   & 16.05\%     & 4.70\%  & 76.99\%   & 18.31\%     \\
                          & W4A4KV    & 91.64\% & 5.86\%    & 2.49\%      & 73.27\% & 18.75\%   & 7.98\%      & 14.62\% & 59.90\%   & 25.48\%     & 4.11\%  & 67.28\%   & 28.62\%     \\
                          & W4A3KV    & 93.04\% & 4.47\%    & 2.50\%      & 76.97\% & 14.78\%   & 8.26\%      & 17.28\% & 53.07\%   & 29.66\%     & 4.96\%  & 60.97\%   & 34.07\%     \\
                          & W3A3KV    & 90.93\% & 5.82\%    & 3.25\%      & 71.48\% & 18.30\%   & 10.23\%     & 13.54\% & 55.46\%   & 31.00\%     & 3.77\%  & 61.73\%   & 34.50\%     \\ \midrule
\multirow{7}{*}{OPT-30b}  & FP16      & 95.12\% & 4.42\%    & 0.46\%      & 82.97\% & 15.42\%   & 1.61\%      & 23.34\% & 69.42\%   & 7.24\%      & 7.07\%  & 84.15\%   & 8.77\%      \\
                          & W4A16     & 82.97\% & 15.42\%   & 1.61\%      & 54.92\% & 40.83\%   & 4.26\%      & 7.07\%  & 84.15\%   & 8.77\%      & 1.87\%  & 88.87\%   & 9.26\%      \\
                          & W4A8      & 90.45\% & 8.41\%    & 1.14\%      & 70.32\% & 26.14\%   & 3.54\%      & 12.90\% & 76.70\%   & 10.40\%     & 3.57\%  & 84.92\%   & 11.51\%     \\
                          & W4A4      & 94.73\% & 4.40\%    & 0.87\%      & 81.79\% & 15.20\%   & 3.01\%      & 21.91\% & 65.17\%   & 12.92\%     & 6.56\%  & 77.98\%   & 15.46\%     \\
                          & W4A4KV    & 94.08\% & 4.37\%    & 1.55\%      & 79.89\% & 14.85\%   & 5.26\%      & 19.89\% & 59.14\%   & 20.97\%     & 5.84\%  & 69.51\%   & 24.64\%     \\
                          & W4A3KV    & 95.14\% & 3.32\%    & 1.54\%      & 83.04\% & 11.57\%   & 5.39\%      & 23.43\% & 52.24\%   & 24.33\%     & 7.10\%  & 63.38\%   & 29.52\%     \\
                          & W3A3KV    & 93.62\% & 4.35\%    & 2.03\%      & 78.59\% & 14.61\%   & 6.80\%      & 18.66\% & 55.49\%   & 25.84\%     & 5.42\%  & 64.53\%   & 30.05\%     \\ \midrule
\multirow{7}{*}{OPT-66b}  & FP16      & 96.21\% & 3.51\%    & 0.27\%      & 86.40\% & 12.62\%   & 0.99\%      & 28.42\% & 66.39\%   & 5.19\%      & 9.03\%  & 84.38\%   & 6.60\%      \\
                          & W4A16     & 86.40\% & 12.62\%   & 0.99\%      & 61.36\% & 35.84\%   & 2.80\%      & 9.03\%  & 84.38\%   & 6.60\%      & 2.42\%  & 90.50\%   & 7.08\%      \\
                          & W4A8      & 92.56\% & 6.76\%    & 0.69\%      & 75.66\% & 22.10\%   & 2.25\%      & 16.27\% & 76.01\%   & 7.73\%      & 4.63\%  & 86.57\%   & 8.80\%      \\
                          & W4A4      & 95.98\% & 3.50\%    & 0.52\%      & 85.64\% & 12.51\%   & 1.86\%      & 27.15\% & 63.42\%   & 9.43\%      & 8.52\%  & 79.64\%   & 11.84\%     \\
                          & W4A4KV    & 95.58\% & 3.49\%    & 0.93\%      & 84.40\% & 12.32\%   & 3.28\%      & 25.27\% & 59.04\%   & 15.70\%     & 7.79\%  & 72.84\%   & 19.37\%     \\
                          & W4A3KV    & 96.44\% & 2.64\%    & 0.92\%      & 87.13\% & 9.54\%    & 3.33\%      & 29.72\% & 52.08\%   & 18.19\%     & 9.56\%  & 67.03\%   & 23.41\%     \\
                          & W3A3KV    & 95.31\% & 3.48\%    & 1.22\%      & 83.54\% & 12.20\%   & 4.26\%      & 24.08\% & 56.27\%   & 19.65\%     & 7.35\%  & 68.67\%   & 23.98\%     \\ \midrule
\multirow{7}{*}{OPT-175b} & FP16      & 97.18\% & 2.68\%    & 0.14\%      & 89.59\% & 9.89\%    & 0.52\%      & 34.98\% & 61.80\%   & 3.22\%      & 11.85\% & 83.78\%   & 4.37\%      \\
                          & W4A16     & 89.59\% & 9.89\%    & 0.52\%      & 68.27\% & 30.16\%   & 1.57\%      & 11.85\% & 83.78\%   & 4.37\%      & 3.25\%  & 91.95\%   & 4.79\%      \\
                          & W4A8      & 94.43\% & 5.21\%    & 0.35\%      & 80.92\% & 17.87\%   & 1.21\%      & 20.95\% & 74.03\%   & 5.02\%      & 6.21\%  & 87.83\%   & 5.95\%      \\
                          & W4A4      & 97.05\% & 2.68\%    & 0.27\%      & 89.18\% & 9.85\%    & 0.98\%      & 33.99\% & 60.06\%   & 5.95\%      & 11.40\% & 80.61\%   & 7.99\%      \\
                          & W4A4KV    & 96.85\% & 2.67\%    & 0.47\%      & 88.49\% & 9.77\%    & 1.73\%      & 32.46\% & 57.37\%   & 10.17\%     & 10.73\% & 75.83\%   & 13.44\%     \\
                          & W4A3KV    & 97.51\% & 2.02\%    & 0.47\%      & 90.73\% & 7.52\%    & 1.75\%      & 37.97\% & 50.32\%   & 11.72\%     & 13.27\% & 70.35\%   & 16.38\%     \\
                          & W3A3KV    & 96.71\% & 2.67\%    & 0.62\%      & 88.02\% & 9.72\%    & 2.26\%      & 31.46\% & 55.59\%   & 12.95\%     & 10.29\% & 72.76\%   & 16.94\%     \\ \bottomrule
\end{tabular}
}
\end{table}

We reported the memory consumption of different parts in LLMs, as shown in the Table~\ref{tab_memory_proportion}.

\end{document}